\documentclass[11pt]{article}

\usepackage[final]{acl}
\usepackage{times}
\usepackage{latexsym}
\usepackage[T1]{fontenc}
\usepackage[utf8]{inputenc}
\usepackage{microtype}
\usepackage{inconsolata}
\usepackage{graphicx}
\usepackage{booktabs}
\usepackage{amssymb}
\usepackage{multirow}
\usepackage{multicol}
\usepackage{pifont}
\usepackage{algorithm}
\usepackage{algorithmic}
\usepackage{bm}
\usepackage[table]{xcolor}
\usepackage[dvipsnames]{xcolor}
\usepackage{colortbl}
\usepackage{subcaption}
\usepackage{caption}
\usepackage{float}
\usepackage{amsmath}
\usepackage{mathtools}
\usepackage{amsthm}
\usepackage{tabularx}
\usepackage{array}
\usepackage{xurl}

\newcolumntype{Y}{>{\raggedright\arraybackslash}X}

\newcommand{\hl}[1]{\cellcolor{gray!13}{#1}}
\newcommand{\cmark}{\textcolor{ForestGreen}{\ding{51}}}
\newcommand{\xmark}{\textcolor{BrickRed}{\ding{55}}}
 
\definecolor{orange}{RGB}{184, 92, 36}
\newcommand{\best}[1]{\cellcolor{orange!15}{\textbf{#1}}}

\newcommand\blfootnote[1]{%
  \begingroup
  \renewcommand\thefootnote{}\footnote{#1}%
  \addtocounter{footnote}{-1}%
  \endgroup
}

\title{Distribution-Consistent Inference for Dynamic Sparse Mixture-of-Experts}

\author{
    Dohyeon Kim$^{1*}$ \quad
    Bedionita Soro$^{1*}$ \quad
    Sung Ju Hwang$^{1,2}$ \\
    $^{1}$ KAIST \quad
    $^{2}$ DeepAuto.ai \\
    \texttt{\{dhkim011030, sorobedio, sjhwang82\}@kaist.ac.kr}
}

\begin{document}
\maketitle

\begin{abstract}
Mixture-of-Experts (MoE) architectures have emerged as a powerful paradigm for scaling model capacity while preserving efficient inference in large foundation models. However, most MoE models use a fixed top-$k$ expert selection policy, assigning the same expert budget to every token even when fewer experts may be sufficient. Inference-time dynamic top-$k$ routing can reduce computation without retraining, but existing methods often overlook the distributional shift caused by deviating from the training-time routing configuration. We show that reducing the number of activated experts consistently increases the RMS scale and variance of SMoE outputs, inducing a representation mismatch that contributes to downstream performance degradation in addition to the loss of expert capacity. To address this correctable component, we propose \textit{Layer-wise Distribution Alignment} (LDA), a lightweight inference-time correction that uses layer-wise calibration statistics to align reduced-routing representations with the default configuration. Across multiple SMoE LLMs, benchmarks, and routing strategies, LDA recovers much of the performance lost induced by the distributional shift under reduced routing while preserving sparse-inference efficiency with negligible overhead.
\end{abstract}

\blfootnote{*Equal contribution.}

\begin{figure}[t]
    \centering
    \begin{subfigure}{\columnwidth}
        \centering
        \includegraphics[width=\columnwidth]{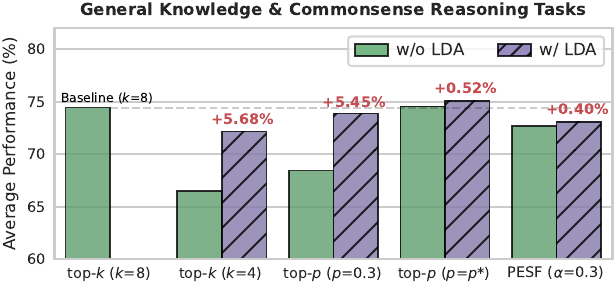}
    \end{subfigure}
    \vspace{1.0em} 
    \begin{subfigure}{\columnwidth}
        \centering
        \includegraphics[width=\columnwidth]{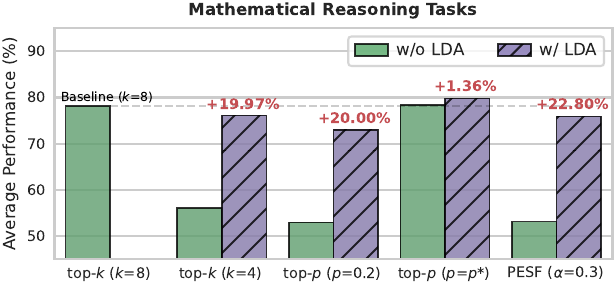}
    \end{subfigure}
    \vspace{-2.5em}
    \caption{\textbf{Comparison of average performance with and without LDA across different routing strategies on Qwen3-30B-A3B.} The dashed line indicates the baseline performance with the top-$k$ routing strategy ($k$=8). $p$* denotes the best-performing top-$p$ threshold for each task, reported in Table~\ref{tab:optimal_top-p}.}
    \label{fig:qwen3_performance}
\end{figure}
\begin{figure*}[t]
    \centering
    \resizebox{\linewidth}{!}{
        \includegraphics[width=\columnwidth]{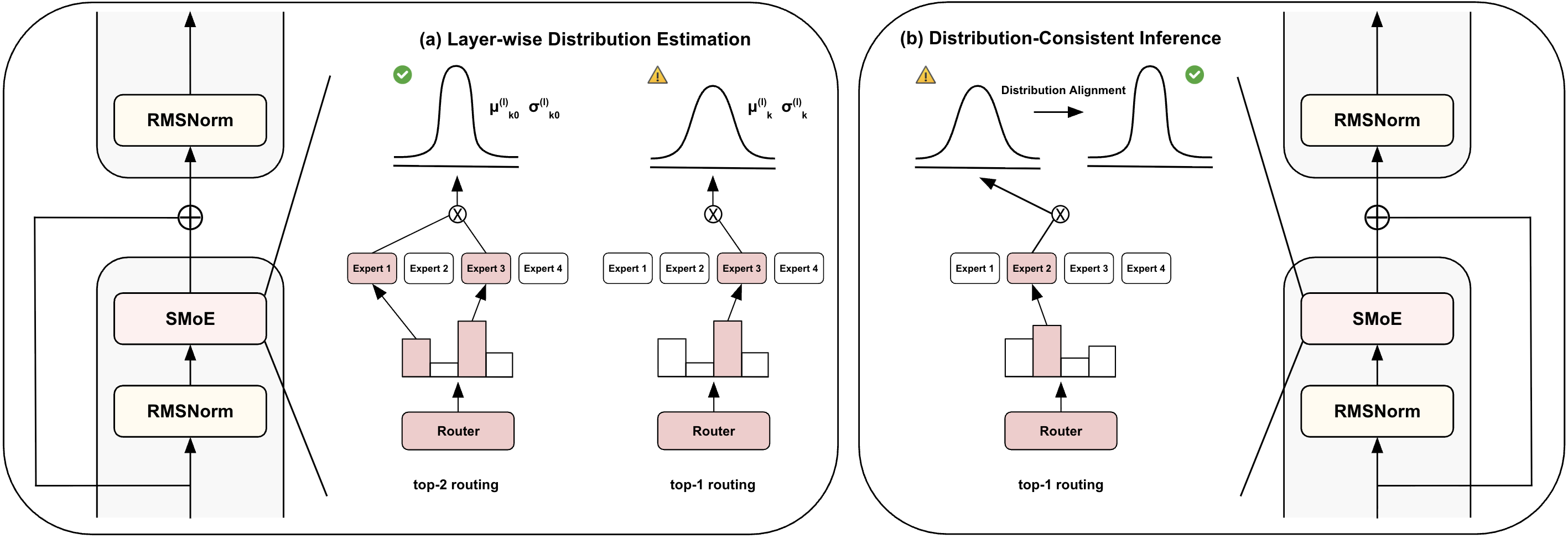}
    }
    \caption{\textbf{Overall framework of Layer-wise Distribution Alignment (LDA).} (a): LDA estimates layer-wise reference and target statistics of SMoE outputs from a calibration dataset. (b): At inference time, LDA aligns reduced-routing SMoE outputs to the reference distribution according to $k$.}
    \label{fig:LDA_framework}
\end{figure*}
\section{Introduction}

The rapid progress of large language models (LLMs) has been driven in large part by scaling model parameters~\citep{llm-survey}, but this trend has also increased computational and memory costs during both training and inference. Sparse Mixture-of-Experts (SMoE) architectures mitigate this bottleneck by decoupling total model capacity from per-token computation~\citep{shazeer2017, moe-survey}, routing each token to only a small subset of experts to expand capacity while maintaining a manageable inference budget. In practice, most SMoE moels employ a fixed top-$k$ routing strategy~\citep{switch_transformer}, where a router scores all experts and selects the $k$ highest-scoring experts for each token. While simple and predictable, this scheme assigns every token the same expert budget regardless of contextual complexity or information density, resulting in unnecessary computation for tokens that could be processed effectively with fewer experts~\citep{topp_baseline}.

Recent work on dynamic expert routing addresses this inefficiency by adapting the number of activated experts per token~\citep{adaptive-gating, zeng-etal-2024-adamoe}, with inference-time variants being particularly attractive because they can be applied directly to pretrained models without retraining~\citep{odp, pesf}. However, existing approaches primarily focus on \emph{when} to reduce expert activation, while largely overlooking a critical consequence: the resulting intermediate representations may deviate from the distribution induced by the training-time routing configuration.

In this work, we systematically analyze how SMoE representations change as a function of the number of activated experts and find that reducing top-$k$ consistently increases the RMS scale and variance of SMoE outputs. This induces a representation-level mismatch that propagates through layers and contributes to downstream performance degradation. These findings suggest that performance loss under reduced routing is not solely a consequence of diminished expert capacity, but also arises from a correctable distributional shift. Motivated by this observation, we propose \textit{Layer-wise Distribution Alignment} (LDA), a lightweight inference-time correction that aligns representations produced under reduced expert activation with the layer-wise reference distribution of the default top-$k_0$ routing configuration. LDA applies a per-dimension moment correction after each SMoE layer, requires no retraining or architectural modification, and naturally complements training-free dynamic routing strategies with negligible computational overhead.

We summarize our contributions as follows:
\begin{itemize}
    \item We show that performance degradation under reduced expert activation is not only due to diminished expert capacity, but is also associated with systematic amplification of the scale and variance of SMoE outputs, a phenomenon not explicitly studied in prior work.
    \item We propose \textit{Layer-wise Distribution Alignment} (LDA), a training-free inference-time correction that mitigates this distributional shift using layer-wise calibration statistics.
    \item We demonstrate that LDA consistently improves the performance--efficiency trade-off across diverse tasks, models, and routing strategies, including top-$k$ routing, adaptive top-$p$ routing, and dynamic expert pruning.
\end{itemize}

\begin{figure*}[t]
    \centering
    \begin{subfigure}[t]{0.32\linewidth}
        \centering
        \includegraphics[width=\columnwidth]{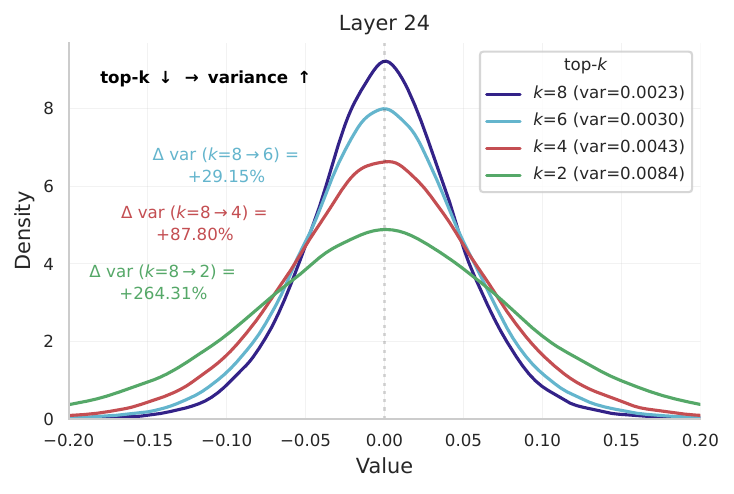}
        \captionsetup{justification=centering}
        \caption{SMoE output distribution}
        \label{fig:qwen3_observations:a}
    \end{subfigure}
    \hfill
    \begin{subfigure}[t]{0.32\linewidth}
        \centering
        \includegraphics[width=\columnwidth]{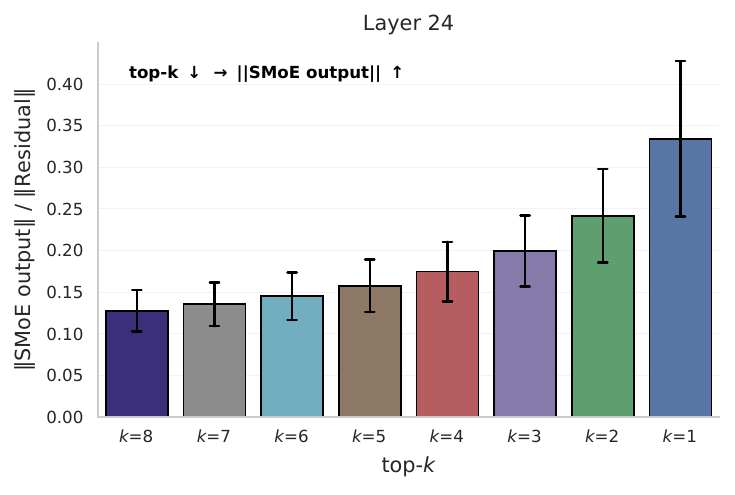}
        \captionsetup{justification=centering}
        \caption{$\|$SMoE output$\|$ / $\|$Residual$\|$}
        \label{fig:qwen3_observations:b}
    \end{subfigure}
    \hfill
    \begin{subfigure}[t]{0.32\linewidth}
        \centering
        \includegraphics[width=\columnwidth]{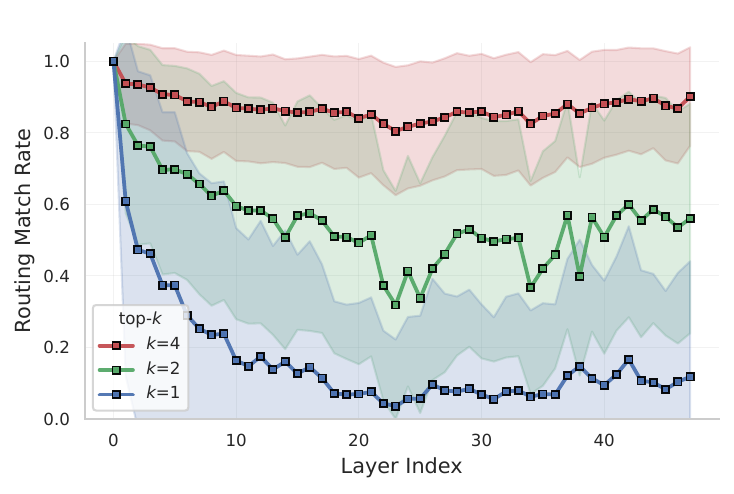} 
        \caption{Layer-wise Routing Match Rate}
        \label{fig:qwen3_observations:c}
    \end{subfigure}
    \caption{\textbf{Observation and analysis of SMoE outputs on Qwen3-30B-A3B.} We use 8$\times$2,048 calibration tokens, and for (a), 100,000 values are randomly sampled. (a): SMoE output distributions become more dispersed as $k$ decreases. (b): The SMoE output scale increases relative to the residual stream as $k$ decreases. (c): The routing match rate (Eq.~\ref{eq:routing_match_rate}) decreases across layers as $k$ becomes smaller, indicating that the routing trajectory increasingly deviates from the default configuration.
    }
    \label{fig:qwen3_observations}
\end{figure*}

\section{Related Works}

Sparse Mixture-of-Experts (SMoE) has been proposed to control computational costs by selectively activating only a small set of experts for each token while greatly expanding the total number of parameters. \citet{shazeer2017} showed that sparse top-$k$ routing is an effective mechanism for scaling model capacity, and this design has since become a standard component of SMoE architectures. Subsequent work extended SMoE to large-scale pretraining~\citep{gshard, switch_transformer}, and several recent LLMs have adopted SMoE architectures~\citep{deepseekmoe, olmoe}.

Most existing SMoE LLMs are pretrained with a fixed top-$k$ routing configuration~\citep{mixtral}, which implicitly shapes the representations learned during training. Prior work improves the performance--efficiency trade-off through dynamic expert allocation based on routing confidence~\citep{topp_baseline}, null experts~\citep{zeng-etal-2024-adamoe}, adaptive top-$k$ selection~\citep{adaptive-gating, AdapMoE}, or training-free expert pruning using token importance~\citep{odp} and expert frequency~\citep{pesf}.

However, these methods primarily focus on routing strategies, leaving underexplored how reducing the number of activated experts affects the distribution of SMoE outputs. In contrast, we address this representation-level mismatch and propose  an efficient training-free correction that aligns representations under reduced expert activation. Our method is complementary to existing training-free dynamic routing strategies.
\section{Observation and Analysis} \label{sec:observation_and_analysis}

\subsection{Preliminaries}
An SMoE layer consists of a router, a gating mechanism, and a set of $N$ experts $\{E_{\theta_i}\}_{i=1}^N$. Given an input $\mathbf{x} \in \mathbb{R}^{D_{\text{in}}}$, the router $G_\theta$ produces logits $G_\theta(\mathbf{x}) \in \mathbb{R}^N$, which are converted into routing scores $\mathbf{w} \in \mathbb{R}^N$ through a gating function, such as softmax~\citep{gshard} or sigmoid~\citep{baselayers_sigmoid_gating}:
\begin{equation*}
    \mathbf{w} =
    \begin{cases}
        \operatorname{softmax}(G_\theta(\mathbf{x})), & \text{(softmax gating)}, \\
        \sigma(G_\theta(\mathbf{x})), & \text{(sigmoid gating)}.
    \end{cases}
\end{equation*}
Under top-$k$ routing, only the $k$ experts with the highest routing scores are selected:
\begin{equation*}
    \mathcal{S}_{k} = \operatorname{TopK}(\mathbf{w}, k),
\end{equation*}
where $\mathcal{S}_{k}$ denotes the set of selected expert indices. Let $\{w_i\}_{i \in \mathcal{S}_{k}}$ denote the routing scores of the selected experts. The selected routing scores can be re-normalized so that they sum to 1:
\begin{equation}
    \label{eq:routing-score_re-normalization}
    w_i =
    \begin{cases}
        \dfrac{w_i}{\sum_{j \in \mathcal{S}_{k}} w_j}, & \text{(re-normalization)}, \\
        w_i, & \text{(otherwise)}.
    \end{cases}
\end{equation}
The output of the SMoE layer is then computed as the weighted sum of the selected expert outputs:
\begin{equation}
    \label{eq:smoe_output}
    \mathbf{y} = \sum_{i \in \mathcal{S}_{k}} w_i E_{\theta_i}(\mathbf{x}),
\end{equation}
where $\mathbf{y} \in \mathbb{R}^{D_{\text{out}}}$ denotes the layer output. 

\subsection{Observation: Reduced top-$k$ Amplifies Representation Variance and Scale} \label{sec:observation}

We focus on SMoE architectures with routing score re-normalization (Eq.~\ref{eq:routing-score_re-normalization}), which is commonly used in recent SMoE LLMs.

\paragraph{Empirical Observation}
We empirically observe that reducing top-$k$ monotonically increases the variance of SMoE outputs. As shown in Fig.~\ref{fig:qwen3_observations:a}, the output distribution becomes more dispersed as fewer experts are activated. Alongside this variance increase, the RMS magnitude of the SMoE output also grows, making it larger relative to the residual stream, as shown in Fig.~\ref{fig:qwen3_observations:b}. These observations suggest that reduced expert activation changes not only the routing sparsity, but also the scale of the resulting representation.

\paragraph{Theoretical Insight} \label{sec:theoretical_insight}

We provide a simple analysis explaining why reducing top-$k$ amplifies the RMS scale of the SMoE output (Eq.~\ref{eq:smoe_output}) under routing score re-normalization, where $\sum_{i \in \mathcal{S}_{k}} w_i = 1$. The RMS is defined as
\begin{equation}
    \label{eq:rms}
    \mathrm{RMS}(\mathbf{y}) = \sqrt{\frac{1}{D}\sum_{i=1}^{D} y_i^2} = \frac{\|\mathbf{y}\|_2}{\sqrt{D}},
\end{equation}
which is determined by the squared norm of $\mathbf{y}$. Expanding this term gives,
\begin{equation}
\label{eq:smoe_output_norm_decomposition}
\begin{aligned}
    \|\mathbf{y}\|_2^2
    &=
    \sum_{i \in \mathcal{S}_{k}} w_i^2 
    \|E_{\theta_i}(\mathbf{x})\|_2^2 \\
    &\quad +
    \sum_{\substack{i,j \in \mathcal{S}_{k} \\ i \neq j}}
    w_i w_j 
    \langle E_{\theta_i}(\mathbf{x}), E_{\theta_j}(\mathbf{x}) \rangle.
\end{aligned}
\end{equation}
Under the mild assumptions that (1) activated expert output scales exhibit limited variation and remain comparable across top-$k$ settings, allowing their squared norms to be approximated by a common scale, and (2) pairwise correlations between expert outputs are weak such that the first diagonal term remains the dominant component,
\begin{equation*}
    \|\mathbf{y}\|_2^2
    \approx
    \sigma^2 \sum_{i \in \mathcal{S}_{k}} w_i^2
    \propto
    \sum_{i \in \mathcal{S}_{k}} w_i^2 ,
\end{equation*}
where $\sigma^2$ denotes the average squared norm of the expert outputs and is treated as a constant.
Thus, the RMS scale of the SMoE output is mainly governed by the concentration of the normalized routing scores. We empirically validate these assumptions in Appendix~\ref{app:theoretical_insight}.

When reducing top-$k$, the remaining routing scores are re-normalized after removing the lower-ranked scores:
\begin{equation*}
    \tilde{w}_i = \frac{w_i}{1-w_k}, \quad i=1,\ldots,k-1 .
\end{equation*}
Let $A = \sum_{i=1}^{k-1} w_i^2$ denote the sum of squared routing scores over the remaining $k-1$ scores, then
% If $A = \sum_{i=1}^{k-1} w_i^2$, then
\begin{equation}
    \label{eq:inequality}
    \sum_{i=1}^{k-1} \tilde{w}_i^2
    =
    \frac{A}{(1-w_k)^2}
    >
    A + w_k^2
    =
    \sum_{i=1}^{k} w_i^2 ,
\end{equation}
where the inequality follows from $w_i \geq w_k$ for $i<k$. 
Therefore, reducing top-$k$ increases the sum of squared normalized scores, which leads to a larger expected squared norm of the SMoE output and explains the RMS amplification. A detailed derivation is provided in Appendix~\ref{app:theoretical_insight}.

\subsection{Structural Analysis}

\paragraph{Modern SMoE LLM Architectures}
Most modern SMoE LLMs adopt residual connections~\citep{resnet} within Pre-LN Transformer architectures~\citep{pre_ln}. In this structure, the SMoE output is added to the residual stream, and the combined representation is normalized in the subsequent block. A commonly used normalization layer is RMSNorm~\citep{rmsnorm}, which rescales a representation by its RMS (Eq.~\ref{eq:rms}):
\begin{equation}
    \label{eq:rms_normalization}
    \mathrm{RMSNorm}(\mathbf{x}) = \boldsymbol{\alpha} \odot \frac{\mathbf{x}}{\mathrm{RMS}(\mathbf{x})} ,
\end{equation}
where $\mathbf{x} \in \mathbb{R}^{D}$ denotes an input representation and $\boldsymbol{\alpha} \in \mathbb{R}^{D}$ denotes a learnable scaling vector. Thus, changes in the scale of the SMoE output can directly affect the scale balance between the existing residual stream and the newly added weighted expert output.

\paragraph{Scale Imbalance between SMoE Output and Residual}
In a residual block, the combined representation can be written as $\mathbf{z} = \mathbf{r} + \mathbf{y}$, where $\mathbf{r} \in \mathbb{R}^{D}$ denotes the residual stream and $\mathbf{y} \in \mathbb{R}^{D}$ denotes the SMoE output. 

The observation above suggests that reducing top-$k$ increases the scale of the SMoE output relative to the residual stream. When the RMS scale of the SMoE output $\mathbf{y}$ increases, the combined representation $\mathbf{z}$ becomes more strongly influenced by the weighted expert output. 

Since RMSNorm rescales the entire representation by a single RMS value, an enlarged SMoE output increases the normalization denominator $\mathrm{RMS}(\mathbf{z})$. As it grows, components carried by the residual stream $\mathbf{r}$ can be relatively suppressed after the normalization layer, even if they contain useful information.

\paragraph{Routing Trajectory Mismatch}
In SMoE models, experts are often specialized to process tokens from different domains or tasks~\citep{domain-specific_expert, interpret_expert}, making it important for each token to be routed to appropriate experts. However, the scale imbalance discussed above can distort the representation passed to subsequent blocks, which may in turn affect the router's expert selection. To quantify this effect, we define \textit{Routing Match Rate} as
\begin{equation}
    \label{eq:routing_match_rate}
    \mathrm{Routing~Match~Rate}
    =
    \frac{
    \left|
    \mathcal{E}_{k}
    \cap
    \mathcal{E}_{k_0}^{(k)}
    \right|
    }{
    \left|
    \mathcal{E}_{k_0}^{(k)}
    \right|
    },
\end{equation}
where $\mathcal{E}_{k}$ denotes the expert set selected under reduced top-$k$ routing, and
$\mathcal{E}_{k_0}^{(k)}$ denotes the top-$k$ subset of experts selected under the default top-$k_0$ routing. A higher routing match rate indicates that reduced top-$k$ routing follows a routing path more similar to the default training configuration.

As shown in Fig.~\ref{fig:qwen3_observations:c}, the routing match rate decreases as the number of activated experts is reduced.
This degradation is especially pronounced for smaller $k$, where the routing path rapidly diverges from the default top-$k_0$ routing across layers. Consequently, these results show that reduced top-$k$ routing also alters the subsequent routing trajectory across layers.
\begin{table}[t]
    \centering
    \resizebox{\columnwidth}{!}{
    \begin{tabular}{ccccc}
    
    \toprule
    \textbf{top-$k$} & \textbf{Mean} & \textbf{Variance} & \textbf{Skewness} & \textbf{Kurtosis (Fisher)} \\

    \midrule
    \multicolumn{1}{c}{} & \multicolumn{4}{c}{Layer 12} \\
    \cmidrule(lr){2-5}
    
    $k$=8 & $0.0000_{\pm 0.0051}$ & $\mathbf{0.0012}_{\pm \mathbf{0.0003}}$ & $-0.0040_{\pm 0.1165}$ & $1.0793_{\pm 0.4076}$ \\
    $k$=4 & $0.0000_{\pm 0.0063}$ & $\mathbf{0.0022}_{\pm \mathbf{0.0005}}$ & $-0.0043_{\pm 0.1140}$ & $1.1111_{\pm 0.4467}$ \\
    $k$=2 & $0.0000_{\pm 0.0073}$ & $\mathbf{0.0044}_{\pm \mathbf{0.0009}}$ & $-0.0060_{\pm 0.1229}$ & $1.3246_{\pm 0.5706}$ \\

    \midrule
    \multicolumn{1}{c}{} & \multicolumn{4}{c}{Layer 24} \\
    \cmidrule(lr){2-5}
    
    $k$=8 & $0.0002_{\pm 0.0060}$ & $\mathbf{0.0022}_{\pm \mathbf{0.0013}}$ & $0.0010_{\pm 0.1110}$ & $0.7520_{\pm 0.6319}$ \\
    $k$=4 & $0.0002_{\pm 0.0069}$ & $\mathbf{0.0042}_{\pm \mathbf{0.0020}}$ & $0.0004_{\pm 0.1055}$ & $0.8080_{\pm 0.6136}$ \\
    $k$=2 & $0.0002_{\pm 0.0078}$ & $\mathbf{0.0083}_{\pm \mathbf{0.0031}}$ & $0.0007_{\pm 0.1193}$ & $1.0866_{\pm 0.9998}$ \\

    \midrule
    \multicolumn{1}{c}{} & \multicolumn{4}{c}{Layer 36} \\
    \cmidrule(lr){2-5}
    
    $k$=8 & $0.0002_{\pm 0.0125}$ & $\mathbf{0.0063}_{\pm \mathbf{0.0056}}$ & $0.0012_{\pm 0.1850}$ & $1.5756_{\pm 1.3413}$ \\
    $k$=4 & $0.0003_{\pm 0.0154}$ & $\mathbf{0.0121}_{\pm \mathbf{0.0096}}$ & $0.0001_{\pm 0.1803}$ & $1.3168_{\pm 1.0057}$ \\
    $k$=2 & $0.0005_{\pm 0.0184}$ & $\mathbf{0.0243}_{\pm \mathbf{0.0156}}$ & $0.0005_{\pm 0.2134}$ & $1.5302_{\pm 1.2054}$ \\
    
    \bottomrule

    \end{tabular}}
    \caption{\textbf{Per-dimension moment statistics of SMoE outputs across different top-$\boldsymbol{k}$ on Qwen3-30B-A3B.} We report $\text{mean}_{\pm \text{std}}$ for each moment. Only the variance consistently shows a substantial increase as $k$ decreases.}
    \label{tab:qwen3_per-dim_statistics}
\end{table}
\begin{figure}[t]
    \centering
    \includegraphics[width=0.9\columnwidth]{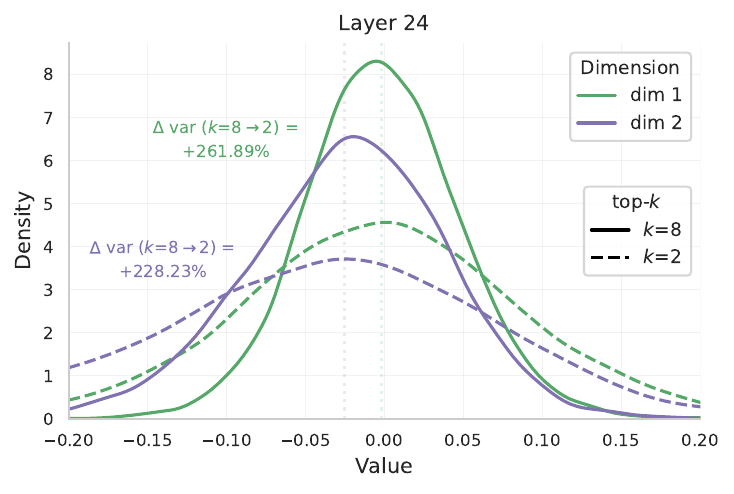}
    \caption{\textbf{Per-dimension distribution of SMoE outputs on Qwen3-30B-A3B.} Each dimension shows different degrees of variance amplification under reduced top-$k$ routing.}
    \label{fig:qwen3_per-dim_variance}
\end{figure}

\section{Layer-wise Distribution Alignment (LDA)}

Motivated by the observation and analysis in Section~\ref{sec:observation_and_analysis}, we use a lightweight inference-time alignment to mitigate the representation scale and variance mismatch induced by reduced top-$k$ routing. The key idea is to use the layer-wise representation statistics under the default top-$k_0$ configuration as a reference, and align the representations produced under reduced top-$k$ routing to the reference distribution. A detailed pseudo-code description is provided in Appendix~\ref{app:algorithms}.

\subsection{Design Choice: Per-dimension Moment Alignment}

\paragraph{Original Representation Space}
Since SMoE outputs computed under top-$k_0$ or different top-$k$ settings lie in the same representation space, we align their statistics in the original representation space instead of applying an additional projection or non-linear transformation.

\paragraph{Per-dimension Alignment}
The analysis in Section~\ref{sec:observation} does not imply that the amplification is uniform across dimensions. As shown in Fig.~\ref{fig:qwen3_per-dim_variance}, the variance shift differs substantially across dimensions, suggesting that global scaling alone does not capture the remaining heterogeneous changes across dimensions. We therefore apply a per-dimension correction, with a direct comparison to global scaling baselines provided in Appendix~\ref{app:simple_scaling_baselines}.

\paragraph{First- and Diagonal Second-order Moment Alignment}
As shown in Table~\ref{tab:qwen3_per-dim_statistics}, the significant and consistent shift appears in the variance, while higher-order moments such as skewness and kurtosis exhibit weaker and less consistent changes. Thus, aligning the per-dimension mean and standard deviation provides a simple affine correction that jointly corrects the amplified scale and the remaining dimension-wise mismatch, while avoiding the computational cost of full-covariance alignment or higher-order transformations.

\subsection{Layer-wise Distribution Estimation}
We estimate the required statistics using a calibration set. 
For each SMoE layer $l$, we compute the per-dimension mean and standard deviation of the SMoE output under the default top-$k_0$ configuration and under reduced top-$k$ routing. 
We denote the reference statistics from the default configuration as
$\boldsymbol{\mu}^{(l)}_{k_0}, \boldsymbol{\sigma}^{(l)}_{k_0} \in \mathbb{R}^{D}$,
and the target statistics from reduced routing as
$\boldsymbol{\mu}^{(l)}_{k}, \boldsymbol{\sigma}^{(l)}_{k} \in \mathbb{R}^{D}$.

During calibration, hidden states are propagated through preceding layers using the default top-$k_0$ routing. 
This prevents distribution shifts from accumulating across layers during statistics estimation and provides a stable reference for measuring how each layer's SMoE output changes under reduced top-$k$ routing.

\begin{table*}[t]
    \centering
    
    \resizebox{\linewidth}{!}{
    \begin{tabular}{l|c|c c|c c c c c c c c c}
        \toprule
        \multicolumn{2}{l}{}
        & \multicolumn{1}{c}{}
        & \multicolumn{1}{c}{}
        & \multicolumn{9}{c}{General Knowledge \& Commonsense Reasoning} \\

        \cmidrule(lr){5-13}
        
        \multicolumn{2}{l}{\textbf{Method}}
        & \multicolumn{1}{c}{\textbf{w/ LDA}}
        & \multicolumn{1}{c}{\textbf{Avg. $k$}}
        & \textbf{MMLU} & \textbf{Hella.} & \textbf{Wino.} & \textbf{ARC-E} & \textbf{ARC-C} & \textbf{CQA} & \textbf{SciQ} & \textbf{PIQA} & \textbf{Average} \\
        
        \midrule
        \multicolumn{2}{l|}{Baseline} & \xmark & 8.00 & \best{77.87\%} & 59.52\% & 70.24\% & 79.80\% & 53.16\% & 79.12\% & 96.50\% & 79.33\% & 74.44\% \\
        
        \cline{1-13}
        \multirow{4}{*}{top-$k$}
        & \multirow{2}{*}{$k$=4}
        & \rule{0pt}{2.6ex}\xmark & 4.00 & 70.72\% & 55.73\% & 60.06\% & 69.57\% & 40.78\% & 65.52\% & 92.80\% & 76.77\% & 66.49\% \\
        & & \hl{\cmark} & \hl{4.00} & \hl{75.56\%} & \hl{57.12\%} & \hl{67.72\%} & \hl{77.44\%} & \hl{47.44\%} & \hl{77.31\%} & \hl{95.90\%} & \hl{78.89\%} & \hl{72.17\%} \\
        
        \cline{2-13}
        & \multirow{2}{*}{$k$=2}
        & \rule{0pt}{2.6ex}\xmark & 2.00 & 27.97\% & 36.02\% & 50.99\% & 41.29\% & 24.83\% & 22.36\% & 70.90\% & 61.48\% & 41.98\% \\
        & & \hl{\cmark} & \hl{2.00} & \hl{67.40\%} & \hl{49.43\%} & \hl{58.41\%} & \hl{71.42\%} & \hl{40.27\%} & \hl{66.17\%} & \hl{93.70\%} & \hl{74.97\%} & \hl{65.22\%} \\

        \cline{1-13}
        \multirow{6}{*}{top-$p$} 
        & \multirow{2}{*}{$p$=0.3}
        & \rule{0pt}{2.6ex}\xmark & 5.62 & 75.35\% & 57.85\% & 66.22\% & 70.33\% & 43.60\% & 60.61\% & 95.00\% & 78.51\% & 68.43\% \\
        & & \hl{\cmark} & \hl{5.43} & \hl{76.46\%} & \hl{58.23\%} & \hl{70.96\%} & \hl{79.38\%} & \hl{51.62\%} & \hl{79.28\%} & \hl{96.00\%} & \hl{79.11\%} & \hl{73.88\%} \\
        
        \cline{2-13}
        & \multirow{2}{*}{$p$=0.2}
        & \rule{0pt}{2.6ex}\xmark & 3.64 & 58.25\% & 51.68\% & 58.25\% & 56.61\% & 34.39\% & 50.45\% & 88.90\% & 72.85\% & 58.92\% \\
        & & \hl{\cmark} & \hl{3.38} & \hl{73.05\%} & \hl{54.65\%} & \hl{64.88\%} & \hl{74.28\%} & \hl{44.20\%} & \hl{76.74\%} & \hl{95.30\%} & \hl{77.26\%} & \hl{70.04\%} \\

        \cline{2-13}
        & \multirow{2}{*}{$p$=$p*$}
        & \rule{0pt}{2.6ex}\xmark & 7.80 & 77.80\% & 59.57\% & 70.72\% & 79.71\% & 53.33\% & 78.95\% & 96.40\% & 79.38\% & 74.55\% \\
        & & \hl{\cmark} & \hl{7.52} & \hl{77.81\%} & \hl{\best{59.75\%}} & \hl{\best{71.98\%}} & \hl{\best{80.26\%}} & \hl{\best{53.92\%}} & \hl{\best{80.51\%}} & \hl{\best{96.70\%}} & \hl{\best{79.71\%}} & \hl{\best{75.08\%}} \\

        \cline{1-13}
        \multicolumn{2}{l|}{\multirow{2}{*}{PESF ($\alpha$=0.3)}}
        & \rule{0pt}{2.6ex}\xmark & 7.87 & 76.20\% & 56.81\% & 68.90\% & 78.11\% & 50.34\% & 77.95\% & 94.80\% & 78.29\% & 72.67\% \\
        \multicolumn{2}{c|}{} & \hl{\cmark} & \hl{7.87} & \hl{76.30\%} & \hl{57.49\%} & \hl{69.53\%} & \hl{79.08\%} & \hl{50.51\%} & \hl{78.29\%} & \hl{94.80\%} & \hl{78.46\%} & \hl{73.07\%} \\

        \bottomrule
    \end{tabular}}

    \vspace{0.5em}
    
    \resizebox{\linewidth}{!}{
    \begin{tabular}{l|c|cc|ccc ccc c}
        \toprule
        \multicolumn{2}{l}{}
        & \multicolumn{1}{c}{}
        & \multicolumn{1}{c}{}
        & \multicolumn{3}{c}{Mathematical Reasoning}
        & \multicolumn{3}{c}{Code Generation}
        & Instruction-Following \\

        \cmidrule(lr){5-7}
        \cmidrule(lr){8-10}
        \cmidrule(lr){11-11}
        
        \multicolumn{2}{l}{\textbf{Method}}
        & \multicolumn{1}{c}{\textbf{w/ LDA}}
        & \multicolumn{1}{c}{\textbf{Avg. $k$}}
        & \textbf{GSM8K} & \textbf{MATH} & \textbf{Average}
        & \textbf{MBPP} & \textbf{HumanEval} & \textbf{Average} 
        & \textbf{IFEval} \\

        \midrule
        \multicolumn{2}{l|}{Baseline} & \xmark & 8.00 & 88.02\% & 68.20\% & 78.11\% & \best{72.60\%} & 84.76\% & 78.68\% & 26.25\% \\
        
        \cline{1-11}
        \multirow{4}{*}{top-$k$}
        & \multirow{2}{*}{$k$=4}
        & \rule{0pt}{2.6ex}\xmark & 4.00 & 74.15\% & 38.00\% & 56.08\% & 34.80\% & 66.46\% & 50.63\% & 24.95\%  \\
        & & \hl{\cmark} & \hl{4.00} & \hl{87.49\%} & \hl{64.60\%} & \hl{76.05\%} & \hl{68.00\%} & \hl{72.56\%} & \hl{70.28\%} & \hl{24.77\%} \\
        
        \cline{2-11}
        & \multirow{2}{*}{$k$=2}
        & \rule{0pt}{2.6ex}\xmark & 2.00 & 00.99\% & 00.00\% & 00.50\% & 00.00\% & 00.61\% & 00.31\% & 09.43\% \\
        & & \hl{\cmark} & \hl{2.00} & \hl{68.54\%} & \hl{42.60\%} & \hl{55.57\%} & \hl{44.40\%} & \hl{36.59\%} & \hl{40.50\%} & \hl{17.74\%} \\

        \cline{1-11}
        \multirow{6}{*}{top-$p$}
        & \multirow{2}{*}{$p$=0.3}
        & \rule{0pt}{2.6ex}\xmark & 7.21 & 86.35\% & 67.40\% & 76.88\% & 67.20\% & 77.44\% & 72.32\% & 28.10\%  \\
        & & \hl{\cmark} & \hl{7.17} & \hl{\best{88.93\%}} & \hl{68.00\%} & \hl{78.47\%} & \hl{71.00\%} & \hl{79.88\%} & \hl{75.44\%} & \hl{\best{30.13\%}} \\
        
        \cline{2-11}
        & \multirow{2}{*}{$p$=0.2}
        & \rule{0pt}{2.6ex}\xmark & 5.46 & 73.16\% & 32.80\% & 52.98\% & 36.00\% & 46.95\% & 41.48\% & 24.03\% \\
        & & \hl{\cmark} & \hl{5.40} & \hl{85.75\%} & \hl{60.20\%} & \hl{72.98\%} & \hl{61.60\%} & \hl{73.17\%} & \hl{67.39\%} & \hl{24.40\%} \\

        \cline{2-11}
        & \multirow{2}{*}{$p$=$p*$}
        & \rule{0pt}{2.6ex}\xmark & 7.87 & 88.02\% & 68.80\% & 78.41\% & 72.40\% & 84.15\% & 78.28\% & 28.84\% \\
        & & \hl{\cmark} & \hl{7.66} & \hl{\best{88.93\%}} & \hl{\best{70.60\%}} & \hl{\best{79.77\%}} & \hl{\best{72.60\%}} & \hl{\best{85.37\%}} & \hl{\best{78.99\%}} & \hl{\best{30.13\%}} \\

        \cline{1-11}
        \multicolumn{2}{l|}{\multirow{2}{*}{PESF ($\alpha$=0.3)}}
        & \rule{0pt}{2.6ex}\xmark & 7.78 & 87.62\% & 18.60\% & 53.11\% & 0.00\% & 76.21\% & 38.11\% & 25.13\% \\
        \multicolumn{2}{c|}{} & \hl{\cmark} & \hl{7.78} & \hl{87.41\%} & \hl{64.40\%} & \hl{75.91\%} & \hl{39.40\%} & \hl{79.26\%} & \hl{59.33\%} & \hl{27.54\%} \\
        
        \bottomrule
    \end{tabular}}
    
    \caption{\textbf{Performance across a wide range of downstream tasks under different routing strategies on Qwen3-30B-A3B.} We use deterministic algorithms for the evaluations. Avg. $k$ denotes the average number of activated experts, and $p$* denotes the best-performing top-$p$ threshold for each task, reported in Table~\ref{tab:optimal_top-p}. Results with the highest performance are highlighted in bold orange.}
    \label{tab:qwen3_performance}
    
\end{table*}

\subsection{Distribution-Consistent Inference}
At inference time, when an SMoE layer uses reduced top-$k$ routing, we align its output to the reference statistics of the corresponding default top-$k_0$ configuration. Given the SMoE output $\mathbf{y}^{(l)} \in \mathbb{R}^{D}$ at layer $l$, LDA applies the following per-dimension affine transformation:
\begin{equation*}
    \hat{\mathbf{y}}^{(l)}
    =
    \begin{cases}
        \mathbf{y}^{(l)}, 
        & (k = k_0), \\[4pt]
        \boldsymbol{\sigma}^{(l)}_{k_0}
        \odot
        \dfrac{
            \mathbf{y}^{(l)} - \boldsymbol{\mu}^{(l)}_{k}
        }{
            \boldsymbol{\sigma}^{(l)}_{k} + \epsilon
        }
        +
        \boldsymbol{\mu}^{(l)}_{k_0},
        & (k < k_0) .
    \end{cases}
\end{equation*}
This operation restores the first- and diagonal second-order statistics of reduced top-$k$ representations toward those of the default training configuration. 
Since it only requires an element-wise affine transformation, LDA introduces negligible inference overhead and requires no additional training or parameter updates. It can also be combined with existing training-free dynamic routing strategies, as it operates independently of how the reduced number of activated experts is selected.
\begin{figure*}[t]
    \centering
    \begin{subfigure}{0.32\linewidth}
        \centering
        \includegraphics[width=\columnwidth]{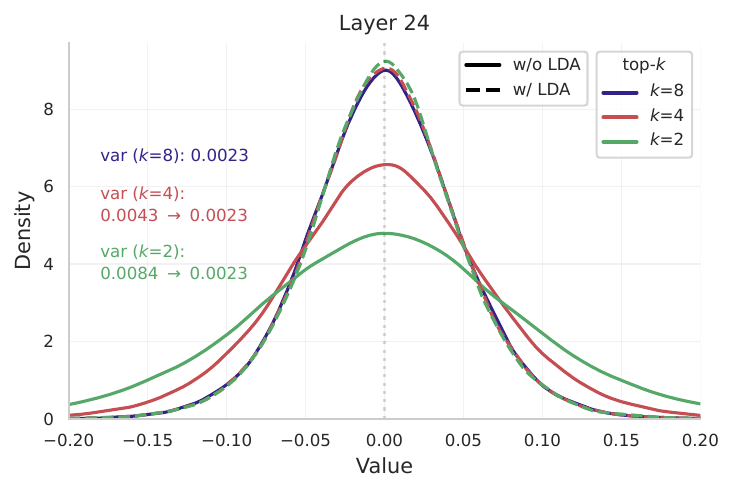}
        \caption{SMoE output distribution}
        \label{fig:qwen3_observations_lda:a}
    \end{subfigure}
    \begin{subfigure}{0.32\linewidth}
        \centering
        \includegraphics[width=\columnwidth]{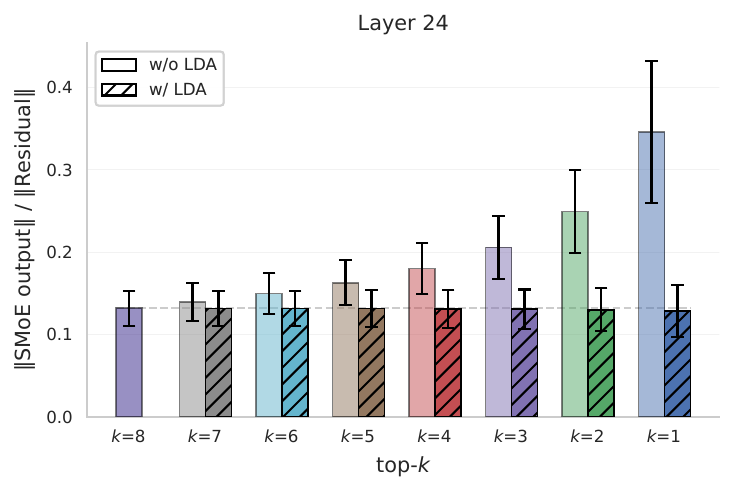}
        \caption{$\|$SMoE output$\|$ / $\|$Residual$\|$}
        \label{fig:qwen3_observations_lda:b}
    \end{subfigure}
    \begin{subfigure}{0.32\linewidth}
        \centering
        \includegraphics[width=\columnwidth]{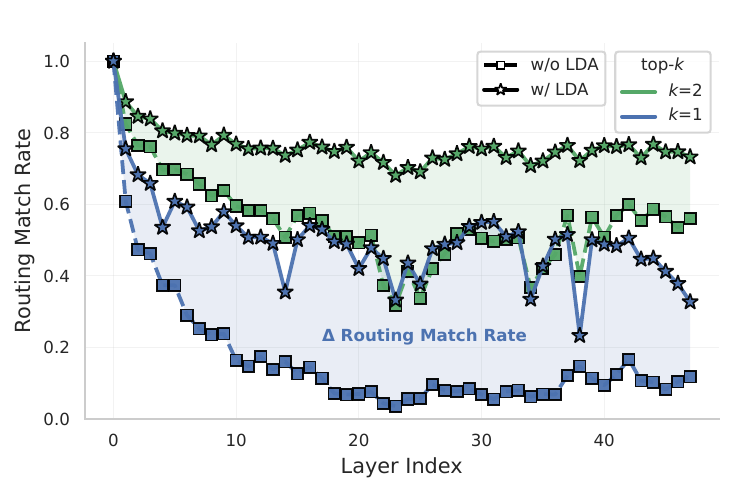}
        \caption{Layer-wise Routing Match Rate}
        \label{fig:qwen3_observations_lda:c}
    \end{subfigure}
    \caption{\textbf{Effect of LDA on SMoE outputs in Qwen3-30B-A3B.} We use 2,048 held-out calibration tokens, and for (a), 100,000 values are randomly sampled. (a): LDA aligns reduced top-$k$ SMoE output distributions toward the default $k=8$ distribution. (b): LDA suppresses the increased SMoE output scale relative to the residual stream. (c) LDA increases the routing match rate (Eq.~\ref{eq:routing_match_rate}) across layers, indicating reduced routing trajectory mismatch.}
    \vspace{-0.5em}
    \label{fig:qwen3_observations_lda}
\end{figure*}

\section{Experiments}

\subsection{Experimental Setup}
We evaluate LDA on recent SMoE LLMs, including Qwen3-30B-A3B~\citep{qwen3}, kanana-2-30b-a3b-instruct~\citep{kanana} based on DeepSeek-V3 architecture~\citep{deepseekv3}, and GLM-4.7-Flash~\citep{glm4.7}. We consider top-$k$ routing, dynamic top-$p$ routing~\citep{topp_baseline}, and PESF~\citep{pesf} as inference-time routing baselines, and compare each routing strategy under reduced expert activation, with and without LDA across 13 benchmarks: 8 general knowledge \& commonsense reasoning tasks, 2 mathematical reasoning tasks, 2 code generation tasks, and 1 instruction-following task. Layer-wise statistics are estimated using 4$\times$2048 tokens sampled from C4~\citep{c4} calibration subset. We report accuracy-based metrics for all benchmark tasks. Additional details on baselines, models, benchmarks, evaluation settings, and implementation are provided in Appendix~\ref{app:experimental_setup}.

\subsection{Performance across Routing Strategies} \label{sec:main_results}

Table~\ref{tab:qwen3_performance} reports results under different inference-time routing strategies, including fixed top-$k$ routing, dynamic top-$p$ routing, and PESF. Across these strategies, performance degradation becomes more pronounced as the average number of activated experts decreases.

Under fixed top-$k$ routing, LDA substantially improves over reduced top-$k$ baselines, particularly in low-$k$ regimes. When $k$ is reduced to 4 or 2, the vanilla baseline suffers large drops across tasks, whereas LDA preserves considerably higher performance under the same expert budget. This suggests that the degradation caused by reducing expert activation is not solely attributable to lower expert capacity, but is also closely tied to the representation variance and scale mismatch analyzed in Section~\ref{sec:observation_and_analysis}.

We observe a similar tendency under dynamic top-$p$ routing and PESF. 
For top-$p$ routing, LDA consistently improves performance at the same threshold, and with an appropriate $p$ value, it can match or exceed the default top-$k_0$ performance while using fewer experts on average. For PESF, LDA also improves performance over the corresponding dynamic expert-pruning baseline, indicating that LDA is effective beyond fixed reduced top-$k$ routing. Overall, these results show that LDA can be combined with diverse training-free dynamic routing strategies by correcting the representation mismatch induced by reduced expert activation.

We provide results on additional models in Appendix~\ref{app:other_performance}, where LDA yields consistent improvements under reduced expert activation budgets.

\begin{table}[t]
    \centering
    \resizebox{\columnwidth}{!}{
    \begin{tabular}{c|cccc}
    
    \toprule
    \multicolumn{1}{c}{\textbf{Task}} & \textbf{top-$k$} & \textbf{w/o LDA} & \textbf{w/ LDA} & \textbf{p-value} \\

    \midrule
    \multirow{2}{*}{MMLU} & $k$=6 & $76.92\%_{\pm 0.00\%}$ & $77.34\%_{\pm 0.00\%}$ & \textbf{0.0341} \\
    & \rule{0pt}{-1.3ex}$k$=4 & $70.72\%_{\pm 0.00\%}$ & $75.56\%_{\pm 0.00\%}$ & \textbf{1.9e-57} \\

    \cline{1-5}
    \multirow{2}{*}{MATH} & \rule{0pt}{2.6ex}$k$=6 & $58.00\%_{\pm 0.89\%}$ & $65.40\%_{\pm 0.49\%}$ & \textbf{1.3e-04} \\
    & \rule{0pt}{-1.3ex}$k$=4 & $35.72\%_{\pm 1.61\%}$ & $61.24\%_{\pm 1.65\%}$ & \textbf{8.4e-05} \\

    \cline{1-5}
    \multirow{2}{*}{MBPP} & \rule{0pt}{2.6ex}$k$=6 & $68.56\%_{\pm 0.64\%}$ & $70.80\%_{\pm 1.29\%}$ & \textbf{0.0308} \\
    & $k$=4 & $43.56\%_{\pm 3.81\%}$ & $68.00\%_{\pm 0.38\%}$ & \textbf{3.7e-04} \\
    
    \bottomrule

    \end{tabular}}
    \caption{\textbf{Statistical significance of LDA improvements.} For MMLU, we apply McNemar's test. For MATH and MBPP, we apply paired $t$-test and report $\text{mean}_{\pm \text{CI95}}$ over 5 random seeds. LDA yields statistically significant improvements ($p<0.05$) across all evaluated settings.}
    \vspace{-0.75em}
    \label{tab:qwen3_statistical_significance}
\end{table}

\subsection{Statistical Significance of Improvements}

To assess whether the observed improvements are attributable to evaluation noise, we conduct statistical significance tests under representative reduced top-$k$ settings. For multiple-choice tasks such as MMLU~\citep{mmlu}, where predictions are deterministic, we apply McNemar's test~\citep{McNemars_test} to paired per-example correctness, comparing reduced top-$k$ routing with and without LDA. For generation tasks such as MATH~\citep{math500} and MBPP~\citep{mbpp}, we evaluate each method across five random seeds and use a paired $t$-test~\citep{paired_t-test} to compare task-level performance. As shown in Table~\ref{tab:qwen3_statistical_significance}, LDA yields statistically significant improvements ($p < 0.05$) across all evaluated settings. We provide the specific null hypotheses and additional testing details in Appendix~\ref{app:statistical_significance_test}.

\begin{table}[t]
    \centering
    \resizebox{\columnwidth}{!}{
    \begin{tabular}{c|cccccc}
        \toprule
         \multicolumn{1}{c}{} & \multicolumn{1}{c}{} & \multicolumn{2}{c}{ShareGPT} & \multicolumn{2}{c}{NuminaMath-1.5} & \multicolumn{1}{c}{} \\
        
        \cmidrule(lr){3-4}
        \cmidrule(lr){5-6}
        
        \multicolumn{1}{c}{\textbf{top-$k$}} & \multicolumn{1}{c}{\textbf{w/ LDA}} & \textbf{TPOT(ms)}$\downarrow$ & \textbf{Throughput(tok/s)}$\uparrow$ & \textbf{TPOT(ms)}$\downarrow$ & \textbf{Throughput(tok/s)}$\uparrow$ & \multicolumn{1}{c}{\textbf{TFLOPs}$\downarrow$} \\

        \midrule
        $k$=8 & \rule{0pt}{-2.6ex}\xmark & 33.22 & 910.71 & 29.97 & 1035.71 & $\approx$ 14.48 \\
        
        \cline{1-7}
        \multirow{2}{*}{$k$=6} & \rule{0pt}{2.6ex}\xmark & 30.14 & 999.78 & 27.01 & 1147.42 & $ \approx$ 12.63 \\
        & \rule{0pt}{-2.6ex}\hl{\cmark} & \hl{30.26} & \hl{995.09} & \hl{26.30} & \hl{1177.90} & \hl{$\approx$ 12.63} \\

        \cline{1-7}
        \multirow{2}{*}{$k$=4} & \rule{0pt}{2.6ex}\xmark & 26.00 & 1156.26 & 22.57 & 1368.72 & $\approx$ 10.77 \\
        & \rule{0pt}{-2.6ex}\hl{\cmark} & \hl{25.97} & \hl{1153.90} & \hl{22.02} & \hl{1402.42} & \hl{$\approx$ 10.77} \\

        \cline{1-7}
        \multirow{2}{*}{$k$=2} & \rule{0pt}{2.6ex}\xmark & 20.11 & 1477.55 & 17.13 & 1780.97 & $\approx$ 8.92 \\
        & \hl{\cmark} & \hl{20.29} & \hl{1461.31} & \hl{16.74} & \hl{1819.19} & \hl{$\approx$ 8.92} \\
        
        \bottomrule
    \end{tabular}}
    \caption{\textbf{Inference efficiency comparison with and without LDA.} We report TPOT, output token throughput, and TFLOPs across different top-$k$ settings on ShareGPT and NuminaMath-1.5 in Qwen3-30B-A3B. TFLOPs are computed with a sequence length of 2,048. The results show that applying LDA introduces negligible additional inference overhead.}
    \label{tab:qwen3_tpot_throughput}

    \vspace{0.5em}
\end{table}
\begin{figure}[h]
    \centering
    \begin{subfigure}[t]{0.49\columnwidth}
        \centering
        \includegraphics[width=\columnwidth]{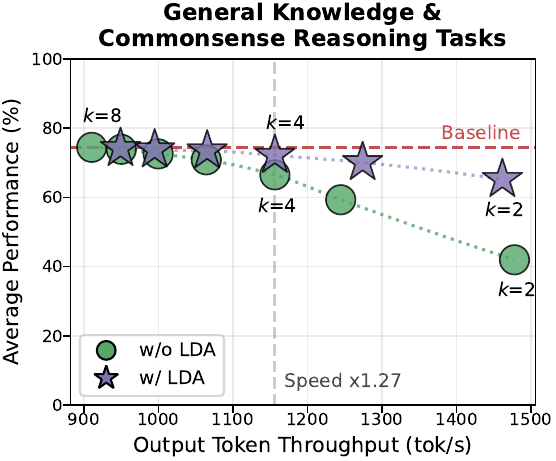}
        \captionsetup{justification=centering}
    \end{subfigure}
    \begin{subfigure}[t]{0.49\columnwidth}
        \centering
        \includegraphics[width=\columnwidth]{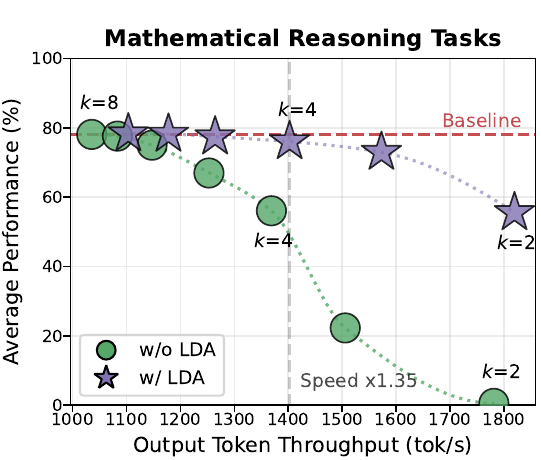}
        \captionsetup{justification=centering}
    \end{subfigure}
    \caption{\textbf{Performance--throughput trade-off on Qwen3-30B-A3B.}
    We compare the average performance and output token throughput under different top-$k$ settings. LDA consistently shifts the trade-off curve upward. Throughput is measured using ShareGPT and NuminaMath-1.5 for general reasoning and mathematical reasoning, respectively.}
    \label{fig:qwen3_throughput}
\end{figure}

\subsection{Effect of LDA}

We analyze how LDA affects representations under reduced top-$k$ routing. As shown in Fig.~\ref{fig:qwen3_observations_lda:a}, LDA aligns the per-dimension distribution of SMoE outputs toward the default top-$k_0$ configuration. After correction, outputs from different reduced top-$k$ settings exhibit comparable means and variances, indicating that LDA mitigates the distributional mismatch identified in Section~\ref{sec:observation}. This correction also stabilizes the relative scale of the SMoE output. Fig.~\ref{fig:qwen3_observations_lda:b} shows that the ratio between the SMoE output scale and the residual stream remains close to that of the default setting after applying LDA, suggesting that LDA reduces the scale imbalance amplified under reduced routing.

We further examine whether this correction affects subsequent routing behavior. Fig.~\ref{fig:qwen3_observations_lda:c} compares the routing match rate in Eq.~\ref{eq:routing_match_rate} before and after LDA. Across layers, LDA consistently increases the match rate relative to the reduced top-$k$ baseline, especially for smaller $k$. This indicates that LDA helps preserve routing trajectories closer to those induced by the default routing configuration.

We observe the same overall tendency on the other evaluated SMoE architectures~\citep{kanana, glm4.7}. These results suggest that the representation-level phenomenon is not confined to a specific architecture despite differences in gating functions, shared-expert structures, and expert configurations. Detailed results are provided in Table~\ref{tab:kanana2_glm4_observations}.

\subsection{Trade-off between Inference Efficiency and Performance}

We evaluate inference efficiency in a realistic serving setting using \texttt{vLLM} framework~\citep{vllm}, measuring \emph{Time Per Output Token} (TPOT) and \emph{Output Token Throughput}. All experiments are conducted on a single NVIDIA RTX PRO 6000 Blackwell Max-Q GPU (96GB). Our analysis focuses on the performance--efficiency trade-off induced by varying the number of activated experts. To capture domain-dependent inference behavior, we use the ShareGPT dataset \citep{vicuna} for general knowledge \& commonsense reasoning, and NuminaMath-1.5 \citep{numina_math} for mathematical reasoning.

Table~\ref{tab:qwen3_tpot_throughput} reports the inference metrics, and Fig.~\ref{fig:qwen3_throughput} visualizes the trade-off between average task performance and output token throughput. As the number of activated experts decreases, inference becomes more efficient, leading to lower FLOPs and higher throughput. However, under vanilla top-$k$ routing, this efficiency gain comes with a substantial performance drop, especially at smaller $k$ values.

In contrast, LDA improves this trade-off by preserving much of the task performance while retaining the efficiency benefit of reduced expert activation. At the same reduced top-$k$ setting, LDA consistently achieves higher average performance than vanilla top-$k$ routing with comparable inference cost. This is because LDA only applies a per-dimension affine transformation to the SMoE output at each layer, whose additional computational cost is $O(LD)$ across the entire model, where $L$ is the number of SMoE layers and $D$ is the hidden dimension. This overhead is negligible compared to the entire computation cost, and it does not require additional training or parameter updates.

This improved trade-off is consistent across both evaluated domains but is especially pronounced in mathematical reasoning, where vanilla reduced top-$k$ routing otherwise suffers catastrophic degradation. Ultimately, these results demonstrate that LDA effectively decouples throughput gains from severe performance penalties, offering a highly practical solution for efficient SMoE serving.

\section{Conclusion}

In this work, we studied the performance degradation of Sparse Mixture-of-Experts (SMoE) models under reduced expert activation from the perspective of representation scale and distribution mismatch. We observed that reducing top-$k$ amplifies the scale and variance of SMoE outputs, disturbing the balance between the weighted expert output and the residual stream. This routing-induced representation mismatch constitutes a correctable component of the degradation, distinct from the loss of expert capacity. To address this issue, we proposed \textit{Layer-wise Distribution Alignment} (LDA), a training-free inference-time correction that aligns reduced top-$k$ representations with the layer-wise statistics of the default training configuration. LDA is lightweight, requiring only an element-wise affine transformation with $O(LD)$ additional cost across the model, and is compatible with training-free dynamic routing strategies. Experiments across multiple SMoE LLMs and diverse downstream tasks show that LDA consistently mitigates performance degradation under reduced expert activation and improves the performance--efficiency trade-off across routing strategies. These results highlight the importance of maintaining representation scale and distribution consistency for dynamic routing in SMoE models.
\section*{Limitations}

LDA is a training-free inference-time correction for the representation scale and distribution mismatch induced by reduced expert activation, and it has several limitations.
First, it corrects the routing-induced mismatch but but like other methods it cannot recover the expert capacity lost by activating fewer experts, so under extreme reduction a capacity gap may remain after alignment and LDA may only partially recover default-routing performance.
Second, LDA summarizes each layer-wise distribution with per-dimension mean and standard deviation. This keeps the method lightweight and stable but ignores covariance structure, asymmetry, and higher-order statistics; although variance is the most consistent and substantial shift under reduced top-$k$ routing, more expressive alignment may help in certain layers or tasks.
Third, LDA requires a calibration stage to estimate layer-wise reference and target statistics. Our ablations show robustness to the choice and size of the calibration set, but the method is not calibration-free.
Finally, our dynamic top-$p$ experiments use manually selected thresholds, which must be chosen per task. Training-free dynamic routing that determines the number of activated experts automatically, combined with LDA, is a useful direction for future work.
\section*{Acknowledgements}

This work was supported by Institute for Information \& communications Technology Planning \& Evaluation (IITP) grant funded by the Korea government(MSIT) (RS-2019-II190075, Artificial Intelligence Graduate School Program (KAIST)), National Research Foundation of Korea (NRF) grant funded by the Korea government (MSIT) (No. RS-2023-00256259), the Institute of Information \& Communications Technology Planning \& Evaluation (IITP) with a grant funded by the Ministry of Science and ICT (MSIT) of the Republic of Korea in connection with the Global AI Frontier Lab International Collaborative Research. (No. RS-2024-00469482 \& RS-2024-00509279), Institute of Information \& communications Technology Planning \& Evaluation(IITP) grant funded by the Korea government(MSIT) (No.RS-2022-II220713, Meta-learning Applicable to Real-world Problems), and the “Advanced GPU Utilization Support Program” funded by the Government of the Republic of Korea (Ministry of Science and ICT).

\bibliography{custom}

\newpage
\appendix

\twocolumn[
\begin{center}
    \begin{minipage}{\textwidth}
    \centering
        \resizebox{\linewidth}{!}{
        \begin{tabular}{c c c ccc ccc ccc}
            \toprule
            \multicolumn{1}{c}{\multirow{2}{*}{\rule[2.4ex]{0pt}{0pt}Layer Index}} 
            & \multicolumn{1}{c}{\multirow{2}{*}{\rule[2.4ex]{0pt}{0pt}$\mathbb{E}_{\mathbf{x}} \left[ \mathrm{PairwiseAbsCos} \right]$}}
            & \multicolumn{1}{c}{\multirow{2}{*}{\rule[2.4ex]{0pt}{0pt}$\mathbb{E}_{\mathbf{x}} \left[ \mathrm{NormCV} \right]$}}
            & \multicolumn{3}{c}{$\mathbb{E}_{\mathbf{x},\, i\in \mathcal{S}_k} \left[ \left\|E_{\theta_i}(\mathbf{x})\right\|_2 \right]$} 
            & \multicolumn{3}{c}{$\mathbb{E}_{\mathbf{x}} \left[ \sum_{i\in \mathcal{S}_k} w_i^2 \right]$} 
            & \multicolumn{3}{c}{$\frac{\mathbb{E}_{\mathbf{x}} \left[ \sum_{\substack{i,j \in \mathcal{S}_k \ i \neq j}} w_i w_j \left\langle E_{\theta_i}(\mathbf{x}), E_{\theta_j}(\mathbf{x})\right\rangle \right]}{ \mathbb{E}_{\mathbf{x}} \left[ \sum_{i \in \mathcal{S}_k} w_i^2 \left\|E_{\theta_i}(\mathbf{x})\right\|_2^2 \right]}$} \\
    
            \cmidrule(lr){4-6}
            \cmidrule(lr){7-9}
            \cmidrule(lr){10-12}
            & & & $k$=8 & $k$=4 & $k$=2 & $k$=8 & $k$=4 & $k$=2 & \;\;\;\; $k$=8 
            \;\;\;\; & \; $k$=4 \; & $k$=2 \\
            
            \midrule
            12 & $0.0321$ & $0.2415$ & $3.2769$ & $3.5664$ & $3.7829$ & $0.1455$ & $0.2675$ & $0.5125$ & $0.1191$ & $0.0726$ & $0.0298$ \\
            24 & $0.0333$ & $0.2256$ & $4.8823$ & $5.2075$ & $5.4284$ & $0.1401$ & $0.2634$ & $0.5102$ & $0.1534$ & $0.0855$ & $0.0326$ \\
            36 & $0.0319$ & $0.2740$ & $7.4705$ & $8.3009$ & $8.9204$ & $0.1440$ & $0.2664$ & $0.5122$ & $0.1427$ & $0.0885$ & $0.0375$ \\
    
            \bottomrule
        \end{tabular}}
    \captionof{table}{\textbf{Empirical validation of the mild assumptions in Qwen3-30B-A3B.} We use 8$\times$2,048 calibration tokens and report the average for each measurement. The results show weak pairwise correlations and limited variation in the norms of activated expert outputs. Routing score concentration increases as $k$ decreases, whereas the cross term remains substantially smaller than the diagonal term, supporting the dominance of the diagonal component.}
    \label{tab:qwen3_empirical_support}
    \end{minipage}
\end{center}
\vspace{1.5em}
]

%%%%%%%%%%%%%%%%%%%%%%%%%%%%%%%%%%%%%%%%
%%%%% Theoretical Insight: Derivation
%%%%%%%%%%%%%%%%%%%%%%%%%%%%%%%%%%%%%%%%
\section{Theoretical Insight} \label{app:theoretical_insight}

We provide additional details for the theoretical insight in Section~\ref{sec:theoretical_insight}. We first derive the inequality~\ref{eq:inequality}. We then provide empirical measurements related to the mild assumptions used in the approximation.

\subsection{Derivation of Eq.~\ref{eq:inequality}}
We derive Eq.~\ref{eq:inequality}, which shows that removing the smallest selected routing score and re-normalizing the remaining scores increases the sum of squared routing scores.

Assume that the selected routing scores are ordered as
\[
w_1 \ge w_2 \ge \cdots \ge w_k > 0,
\]
and normalized such that
\[
\sum_{i=1}^{k} w_i = 1.
\]
After removing the smallest selected score $w_k$, the remaining $k-1$ scores are re-normalized as
\[
\tilde{w}_i = \frac{w_i}{1-w_k},
\quad i=1,\ldots,k-1.
\]
Let $A$ denote the sum of squared routing scores over the remaining $k-1$ scores before re-normalization,
\[
A = \sum_{i=1}^{k-1} w_i^2.
\]
Then,
\[
\sum_{i=1}^{k-1} \tilde{w}_i^2
=
\sum_{i=1}^{k-1}
\left(\frac{w_i}{1-w_k}\right)^2
=
\frac{A}{(1-w_k)^2}.
\]

We now show that this quantity is larger than the original squared sum over all $k$ selected scores,
\[
\frac{A}{(1-w_k)^2} > A+w_k^2.
\]
Multiplying both sides by $(1-w_k)^2>0$,
\[
A > (A+w_k^2)(1-w_k)^2.
\]
Rearranging the inequality gives
\[
A\,w_k(2-w_k)
>
\left(w_k(1-w_k)\right)^2.
\]
Since $w_i \ge w_k$ for all $i<k$,
\[
A
=
\sum_{i=1}^{k-1} w_i^2
\ge
w_k \sum_{i=1}^{k-1} w_i
=
w_k(1-w_k).
\]
Using $A \ge w_k(1-w_k)$ and $0 < w_k < 1$,
\[
A\,w_k(2-w_k)
\ge
w_k(1-w_k)\,w_k(2-w_k).
\]
Moreover, since $2-w_k > 1-w_k$,
\[
w_k(1-w_k)\,w_k(2-w_k)
>
\left(w_k(1-w_k)\right)^2.
\]
Thus,
\[
A\,w_k(2-w_k)
>
\left(w_k(1-w_k)\right)^2,
\]
which proves
\[
\frac{A}{(1-w_k)^2} > A+w_k^2.
\]
This shows that removing the smallest routing score and re-normalizing the remaining scores increases the sum of squared routing scores. 

\subsection{Empirical Support for Mild Assumptions}
The theoretical insight in Section~\ref{sec:theoretical_insight} relies on the mild assumptions that (1) activated expert output scales exhibit limited variation and remain comparable across top-$k$ settings, and (2) pairwise correlations between expert outputs are weak such that the first diagonal term remains the dominant component. To empirically examine these assumptions, we report statistics based on activated expert outputs and normalized routing scores in Table~\ref{tab:qwen3_empirical_support}.

We first measure the average absolute pairwise cosine similarity between activated expert outputs under the default routing configuration $k_0$. We refer to this quantity as $\mathrm{PairwiseAbsCos}$:
\[
\mathbb{E}_{\substack{
i,j \in \mathcal{S}_{k_0}\\
i \neq j
}}
\left[
\left|
\cos\left(
E_{\theta_i}(\mathbf{x}),
E_{\theta_j}(\mathbf{x})
\right)
\right|
\right].
\]
We further quantify the relative variation in the norms of activated expert outputs using their coefficient of variation, which we refer to as $\mathrm{NormCV}$:
\[
\frac{
\sigma_{i \in \mathcal{S}_{k_0}}
\left(
\left\|E_{\theta_i}(\mathbf{x})\right\|_2
\right)
}{
\mu_{i \in \mathcal{S}_{k_0}}
\left(
\left\|E_{\theta_i}(\mathbf{x})\right\|_2
\right)
}.
\]
We then measure the average norm of activated expert outputs and the sum of squared normalized routing scores under different top-$k$ settings:
\[
\mathbb{E}_{i \in \mathcal{S}_k}
\left[
\left\|E_{\theta_i}(\mathbf{x})\right\|_2
\right],
\;
\sum_{i\in \mathcal{S}_k} w_i^2.
\]

As shown in Table~\ref{tab:qwen3_empirical_support}, the average norm of activated expert outputs remains within a comparable range, supporting the assumption that their scale does not vary substantially. In contrast, the sum of squared normalized routing scores increases substantially as $k$ decreases, indicating increased concentration of the normalized routing scores.

Finally, we directly compare the second cross term with the first diagonal term in Eq.~\ref{eq:smoe_output_norm_decomposition}. Across all evaluated layers and top-$k$ setting, the cross term remains substantially smaller than the diagonal term, and its relative contribution further decreases as $k$ decreases. This supports the assumption that the first diagonal term remains the dominant component. Together, these results provide empirical support for the approximation used in Section~\ref{sec:theoretical_insight} and the RMS amplification mechanism under reduced top-$k$ routing.

\begin{table*}[t]
    \centering
    \resizebox{\linewidth}{!}{\begin{tabular}{l c c c c}
    
    \toprule
    \multicolumn{1}{l}{\textbf{Task}} & \textbf{Metric} & \textbf{Few-shot} & \textbf{Max Seq. Length} & \textbf{Max Gen. Tokens} \\

    \midrule
    \multicolumn{1}{l}{\textbf{General Knowledge \& Commonsense Reasoning}} & & & & \\
    \rule{0pt}{2.0ex}MMLU~\citep{mmlu} & Accuracy & 0-shot & 4096 & - \\
    Hellaswag~\citep{hellaswag} & Accuracy & 0-shot & 4096 & - \\
    Winogrande~\citep{winogrande} & Accuracy & 0-shot & 4096 & - \\
    ARC-Easy~\citep{arc} & Accuracy & 0-shot & 4096 & - \\
    ARC-Challenge~\citep{arc} & Accuracy & 0-shot & 4096 & - \\
    CommonsenseQA~\citep{commonsense_qa} & Accuracy & 0-shot & 4096 & - \\
    ScienceQA~\citep{sciq} & Accuracy & 0-shot & 4096 & - \\
    PIQA~\citep{piqa} & Accuracy & 0-shot & 4096 & - \\
    
    \midrule
    \multicolumn{1}{l}{\textbf{Mathematical Reasoning}} \\
    \rule{0pt}{2.0ex}GSM8K~\citep{gsm8k} & Exact Match & 8-shot (single-turn) & 4096 & 1024 \\
    MATH~\citep{math500} & Exact Match & 4-shot (single-turn) & 4096 & 1024 \\

    \midrule
    \multicolumn{1}{l}{\textbf{Code Generation}} \\
    \rule{0pt}{2.0ex}MBPP~\citep{mbpp} & Pass@1 & 3-shot (single-turn) & 4096 & 1024 \\
    HumanEval~\citep{humaneval} & Pass@1 & 0-shot & 4096 & 1024 \\
    
    \midrule
    \multicolumn{1}{l}{\textbf{Instruction-Following}} \\
    \rule{0pt}{2.0ex}IFEval~\citep{ifeval} & Accuracy[prompt-level] & 0-shot & 4096 & 1024 \\

    \bottomrule
    \end{tabular}}

    \caption{\textbf{Benchmark-specific evaluation settings.} We report the evaluation metric, few-shot, max sequence length, and max generation tokens for each benchmark.}
    \label{tab:benchmark_settings}
\end{table*}

%%%%%%%%%%%%%%%%%%%%%%%%%%%%%%%%%%%%%%%%
%%%%% Experimental Setup
%%%%%%%%%%%%%%%%%%%%%%%%%%%%%%%%%%%%%%%%
\section{Experimental Setup} \label{app:experimental_setup}

We provide further details on our experimental setup, including baselines, models, benchmarks, and evaluation protocols.

\subsection{Baselines}
We consider three routing strategies as baselines: fixed top-$k$ routing, dynamic top-$p$ routing~\citep{topp_baseline} and PESF~\citep{pesf}. For each routing strategy, we compare reduced expert activation with and without LDA to validate the effect of distribution alignment.

\paragraph{top-$\boldsymbol{k}$ Routing}
top-$k$ routing is the standard routing strategy used in many SMoE LLMs, where each token activates a fixed number of experts. This setting provides a controlled comparison of LDA by keeping the expert budget identical between the vanilla reduced top-$k$ baseline and its LDA-applied counterpart.

\paragraph{top-$\boldsymbol{p}$ Routing}
top-$p$ routing is a dynamic expert routing strategy that selects experts until the cumulative routing score exceeds a threshold $p$. Unlike fixed top-$k$ routing, the number of activated experts can vary across tokens depending on the routing distribution. To preserve the inference--efficiency motivation, we restrict the maximum number of activated experts so that it does not exceed the default top-$k_0$ configuration. This setting allows us to evaluate whether LDA remains effective when the number of activated experts is determined dynamically at inference time. The evaluated threshold values are described in Appendix~\ref{app:evaluation_protocols}.

\paragraph{PESF}
Pruning based on Expert-Selection Frequency (PESF) is a dynamic expert pruning method applied at inference time. For each input sequence, PESF counts how many times each expert is selected by the router. Then, An expert is pruned if its selection count is smaller than a predefined threshold:
\[
c_i < \frac{l \times K}{N} \times \alpha,
\]
where $c_i$ denotes the number of times expert $E_{\theta_i}$ is selected, $l$ is the input sequence length, $K$ is the number of selected experts per token before pruning, $N$ is the total number of experts, and $\alpha$ is the pruning threshold. Since the set of retained experts can vary across each token, PESF can be viewed as an inference-time dynamic routing strategy that changes the number of activated experts. We set $\alpha \in \{0.3, 0.7\}$ following prior work.

\subsection{Models} \label{app:models}
We evaluate LDA on recent SMoE LLMs with diverse architectures to verify that it is not limited to a specific model structure and works consistently across different architectural designs. All evaluated models apply routing score re-normalization (Eq.~\ref{eq:routing-score_re-normalization}).

\paragraph{Qwen3-30B-A3B~\citep{qwen3}}
This model has 30.5B total parameters, with 3.3B activated parameters per token. Each MoE layer consists of 128 experts, among which 8 routed experts are activated per token. The routing scores are computed using softmax gating.

\paragraph{kanana-2-30b-a3b-instruct~\citep{kanana}}
This model is based on the DeepSeek-V3 architecture~\citep{deepseekv3} and has 30B total parameters with approximately 3B activated parameters per token. Each MoE layer consists of 128 experts, among which 8 experts (2 shared experts and 6 routed experts) are activated per token. The routing scores are computed using sigmoid gating.

\paragraph{GLM-4.7-Flash~\citep{glm4.7}}
This model has 30B total parameters with approximately 3B activated parameters per token. Each MoE layer consists of 64 experts, among which 5 experts (1 shared expert + 4 routed experts) are activated per token. The routing scores are calculated with sigmoid gating.

\subsection{Benchmarks} \label{app:benchmarks}
We evaluate LDA across four domains of benchmarks: general knowledge
\& commonsense reasoning, mathematical reasoning, code generation, and
instruction-following.

\paragraph{General Knowledge \& Commonsense Reasoning Tasks:} 
MMLU \citep{mmlu}, Hellaswag \citep{hellaswag}, Winogrande \citep{winogrande}, ARC (ARC-Easy, ARC-Challenge) \citep{arc}, CommonsenseQA (CQA) \citep{commonsense_qa}, ScienceQA (SciQ) \citep{sciq}, and PIQA \citep{piqa}.

\paragraph{Mathematical Reasoning Tasks:} 
GSM8K \citep{gsm8k} and MATH \citep{math500}.

\paragraph{Code Generation Tasks:} 
MBPP \citep{mbpp} and HumanEval \citep{humaneval}.

\paragraph{Instruction-Following Tasks:} 
IFEval \citep{ifeval}.

\begin{table*}[t]
    \centering
    
    \resizebox{\linewidth}{!}{
    \begin{tabular}{l|c|c c|c c c c c c c c c}
        \toprule
        \multicolumn{2}{l}{}
        &
        & \multicolumn{1}{c}{}
        & \multicolumn{9}{c}{General Knowledge \& Commonsense Reasoning} \\

        \cmidrule(lr){5-13}
        
        \multicolumn{2}{l}{\textbf{Method}}
        & \textbf{w/ LDA} 
        & \multicolumn{1}{c}{\textbf{Avg. $k$}}
        & \rule{0pt}{2.6ex}\textbf{MMLU} & \textbf{Hella.} & \textbf{Wino.} & \textbf{ARC-E} & \textbf{ARC-C} & \textbf{CQA} & \textbf{SciQ} & \textbf{PIQA} & \textbf{Average} \\
        
        \midrule
        
        \multicolumn{2}{l|}{\rule[-1.2ex]{0pt}{3.0ex}Baseline} & \xmark & 2+6.00 &73.63\% & \best{62.16\%} & 72.14\% & 85.69\% & 58.79\% & 65.68\% & 97.00\% & 80.69\% & 74.47\% \\
        
        \cline{1-13}
        \multirow{6}{*}{top-$k$}
        & \multirow{2}{*}{$k$=2+4}
        & \rule{0pt}{2.6ex}\xmark & 2+4.00 & 71.81\% & 61.13\% & 71.27\% & 84.43\% & 55.20\% & 59.95\% & 96.50\% & 79.49\% & 72.47\% \\
        & & \hl{\cmark} & \hl{2+4.00} & \hl{73.33\%} & \hl{61.31\%} & \hl{72.69\%} & \hl{86.03\%} & \hl{56.57\%} & \hl{65.85\%} & \hl{97.20\%} & \hl{80.52\%} & \hl{74.19\%} \\
        
        \cline{2-13}
        & \multirow{2}{*}{$k$=2+2}
        & \rule{0pt}{2.6ex}\xmark & 2+2.00 & 60.87\% & 53.76\% & 58.96\% & 74.28\% & 45.56\% & 46.36\% & 90.20\% & 75.08\% & 63.13\% \\
        & & \hl{\cmark} & \hl{2+2.00} & \hl{68.65\%} & \hl{55.09\%} & \hl{65.82\%} & \hl{83.92\%} & \hl{53.50\%} & \hl{63.31\%} & \hl{96.20\%} & \hl{78.18\%} & \hl{70.58\%} \\
        
        \cline{2-13}
        & \multirow{2}{*}{$k$=2+1}
        & \rule{0pt}{2.6ex}\xmark & 2+1.00 & 23.89\% & 32.28\% & 51.30\% & 35.27\% & 20.56\% & 19.41\% & 54.40\% & 56.09\% & 36.65\% \\
        & & \hl{\cmark} & \hl{2+1.00} & \hl{49.42\%} & \hl{44.04\%} & \hl{56.12\%} & \hl{70.62\%} & \hl{37.80\%} & \hl{43.41\%} & \hl{92.90\%} & \hl{70.46\%} & \hl{58.10\%} \\

        \cline{1-13}
        \multirow{6}{*}{top-$p$} 
        & \multirow{2}{*}{$p$=0.2}
        & \rule{0pt}{2.6ex}\xmark & 2+3.45 & 69.21\% & 59.50\% & 66.61\% & 81.19\% & 50.60\% & 57.08\% & 95.00\% & 77.86\% & 69.63\% \\
        & & \hl{\cmark} & \hl{2+3.52} & \hl{70.43\%} & \hl{59.50\%} & \hl{69.38\%} & \hl{84.72\%} & \hl{55.89\%} & \hl{67.24\%} & \hl{96.70\%} & \hl{77.97\%} & \hl{72.73\%} \\
        
        \cline{2-13}
        & \multirow{2}{*}{$p$=0.1}
        & \rule{0pt}{2.6ex}\xmark & 2+2.50 & 23.66\% & 31.72\% & 50.12\% & 35.27\% & 20.22\% & 19.82\% & 42.00\% & 59.09\% & 35.24\% \\
        & & \hl{\cmark} & \hl{2+1.93} & \hl{58.92\%} & \hl{49.05\%} & \hl{60.77\%} & \hl{78.49\%} & \hl{45.99\%} & \hl{58.48\%} & \hl{94.60\%} & \hl{73.50\%} & \hl{64.98\%} \\

        \cline{2-13}
        & \multirow{2}{*}{$p$=$p*$}
        & \rule{0pt}{2.6ex}\xmark & 2+5.93 & 73.62\% & 62.01\% & 72.38\% & 85.69\% & 59.13\% & 65.60\% & 97.00\% & 80.63\% & 74.50\% \\
        & & \hl{\cmark} & \hl{2+5.75} & \hl{\best{73.91\%}} & \hl{62.06\%} & \hl{\best{73.80\%}} & \hl{\best{85.88\%}} & \hl{\best{59.48\%}} & \hl{\best{66.18\%}} & \hl{\best{97.30\%}} & \hl{\best{80.90\%}} & \hl{\best{74.94\%}} \\

        \cline{1-13}
        \multicolumn{2}{c|}{\multirow{2}{*}{PESF ($\alpha$=0.3)}}
        & \rule{0pt}{2.6ex}\xmark & 2+5.94 & 73.03\% & 61.09\% & 72.16\% & 84.93\% & 57.32\% & 64.78\% & 97.00\% & 79.70\% & 73.76\% \\
        \multicolumn{2}{c|}{} & \hl{\cmark} & \hl{2+5.94} & \hl{73.23\%} & \hl{61.09\%} & \hl{72.69\%} & \hl{85.05\%} & \hl{57.93\%} & \hl{65.11\%} & \hl{97.00\%} & \hl{79.81\%} & \hl{73.99\%} \\

        \bottomrule
    \end{tabular}}

    \vspace{0.5em}
    
    \resizebox{\linewidth}{!}{
    \begin{tabular}{l c c c c c c c c c c}
        \toprule
        \multicolumn{2}{l}{}
        & \multirow{2}{*}{}
        & \multirow{2}{*}{}
        & \multicolumn{3}{c}{Mathematical Reasoning}
        & \multicolumn{3}{c}{Code Generation}
        & Instruction-Following \\

        \cmidrule(lr){5-7}
        \cmidrule(lr){8-10}
        \cmidrule(lr){11-11}
        
        \multicolumn{2}{l}{\textbf{Method}}
        & \textbf{w/ LDA}
        & \textbf{Avg. $k$}
        & \textbf{GSM8K} & \textbf{MATH} & \textbf{Average}
        & \textbf{MBPP} & \textbf{HumanEval} & \textbf{Average} 
        & \textbf{IFEval} \\

        \midrule
        \multicolumn{2}{l|}{\rule[-1.2ex]{0pt}{3.0ex}Baseline} & \xmark & \multicolumn{1}{c|}{2+6.00} & 87.95\% & 64.60\% & 76.28\% & 69.80\% & 82.93\% & 76.37\% & 42.33\% \\
        
        \cline{1-11}
        \multicolumn{1}{l|}{\multirow{6}{*}{top-$k$}}
        & \multicolumn{1}{c|}{\multirow{2}{*}{$k$=2+4}}
        & \rule{0pt}{2.6ex}\xmark & \multicolumn{1}{c|}{2+4.00} & 83.78\% & 55.80\% & 69.79\% & 65.40\% & 81.32\% & 73.36\% & 41.04\% \\
        \multicolumn{1}{l|}{} & \multicolumn{1}{c|}{} & \hl{\cmark} & \multicolumn{1}{c|}{\hl{2+4.00}} & \hl{86.81\%} & \hl{66.40\%} & \hl{76.61\%} & \hl{68.20\%} & \hl{81.71\%} & \hl{74.96\%} & \hl{40.30\%} \\
        
        \cline{2-11}
        \multicolumn{1}{l|}{} & \multicolumn{1}{c|}{\multirow{2}{*}{$k$=2+2}} & \rule{0pt}{2.6ex}\xmark & \multicolumn{1}{c|}{2+2.00} & 49.66\% & 10.80\% & 30.23\% & 17.40\% & 37.80\% & 27.60\% & 25.14\%  \\
        \multicolumn{1}{l|}{} & \multicolumn{1}{c|}{} & \hl{\cmark} & \multicolumn{1}{c|}{\hl{2+2.00}} & \hl{79.53\%} & \hl{55.60\%} & \hl{67.57\%} & \hl{62.40\%} & \hl{82.93\%} & \hl{72.67\%} & \hl{38.82\%} \\
        
        \cline{2-11}
        \multicolumn{1}{l|}{} & \multicolumn{1}{c|}{\multirow{2}{*}{$k$=2+1}}
        & \rule{0pt}{2.6ex}\xmark & \multicolumn{1}{c|}{2+1.00} & 00.23\% & 00.00\% & 00.12\% & 00.00\% & 00.00\% & 00.00\% & 07.95\% \\
        \multicolumn{1}{l|}{} & \multicolumn{1}{c|}{} & \hl{\cmark} & \multicolumn{1}{c|}{\hl{2+1.00}} & \hl{33.97\%} & \hl{13.00\%} & \hl{23.49\%} & \hl{29.40\%} & \hl{32.32\%} & \hl{30.86\%} & \hl{23.66\%} \\

        \cline{1-11}
        \multicolumn{1}{l|}{\multirow{6}{*}{top-$p$}}
        & \multicolumn{1}{c|}{\multirow{2}{*}{$p$=0.2}}
        & \rule{0pt}{2.6ex}\xmark & \multicolumn{1}{c|}{2+4.41} & 79.30\% & 46.60\% & 62.95\% & 54.60\% & 73.78\% & 64.19\% & 38.82\% \\
        \multicolumn{1}{l|}{} & \multicolumn{1}{c|}{} & \hl{\cmark} & \multicolumn{1}{c|}{\hl{2+4.78}} & \hl{81.73\%} & \hl{59.40\%} & \hl{70.57\%} & \hl{63.20\%} & \hl{\best{84.15\%}} & \hl{73.68\%} & \hl{41.77\%} \\
        
        \cline{2-11}
        \multicolumn{1}{l|}{} & \multicolumn{1}{c|}{\multirow{2}{*}{$p$=0.1}}
        & \rule{0pt}{2.6ex}\xmark & \multicolumn{1}{c|}{2+3.72} & 00.30\% & 00.00\% & 00.15\% & 00.00\% & 00.00\% & 00.00\% & 14.60\% \\
        \multicolumn{1}{l|}{} & \multicolumn{1}{c|}{} & \hl{\cmark} & \multicolumn{1}{c|}{\hl{2+3.74}} & \hl{59.82\%} & \hl{38.20\%} & \hl{49.01\%} & \hl{46.60\%} & \hl{61.59\%} & \hl{54.10\%} & \hl{31.79\%} \\

        \cline{2-11}
        \multicolumn{1}{l|}{} & \multicolumn{1}{c|}{\multirow{2}{*}{$p$=$p*$}}
        & \rule{0pt}{2.6ex}\xmark & \multicolumn{1}{c|}{2+5.94} & 87.79\% & 64.00\% & 75.90\% & 69.20\% & 82.93\% & 76.07\% & 44.73\%\\
        \multicolumn{1}{l|}{} & \multicolumn{1}{c|}{} & \hl{\cmark} & \multicolumn{1}{c|}{\hl{2+5.80}} & \hl{\best{88.48\%}} & \hl{\best{66.20\%}} & \hl{\best{77.34\%}} & \hl{\best{70.20\%}} & \hl{\best{84.15\%}} & \hl{\best{77.18\%}} & \hl{\best{46.58\%}} \\

        \cline{1-11}
        \multicolumn{2}{l|}{\multirow{2}{*}{PESF ($\alpha$=0.3)}}
        & \rule{0pt}{2.6ex}\xmark & \multicolumn{1}{c|}{2+5.79} & 86.80\% & 61.20\% & 74.00\% & 66.20\% & 84.14\% & 75.17\% & 43.99\% \\
        \multicolumn{2}{l|}{} & \hl{\cmark} & \multicolumn{1}{c|}{\hl{2+5.79}} & \hl{87.03\%} & \hl{63.40\%} & \hl{75.30\%} & \hl{67.00\%} & \hl{85.36\%} & \hl{76.18\%} & \hl{44.91\%} \\
        \bottomrule
    \end{tabular}}
    
    \caption{\textbf{Performance across a wide range of downstream tasks under different routing strategies on kanana-2-30b-a3b-instruct.} We use deterministic algorithms for the evaluations. Avg. $k$ denotes the average number of activated experts, and $p$* denotes the best-performing top-$p$ threshold for each task, reported in Table~\ref{tab:optimal_top-p}. Results with the highest performance are highlighted in bold orange.}
    \label{tab:kanana2_performance}
\end{table*}

\subsection{Evaluation Protocols} \label{app:evaluation_protocols}
We use \texttt{lm\_eval} framework~\citep{eval-harness} for benchmark evaluations. For reproducibility, we use the default random seed configuration provided by the evaluation framework. We categorize the evaluated benchmarks into option-based tasks and generation-based tasks, and detailed benchmark-specific settings are provided in Table~\ref{tab:benchmark_settings}.

\paragraph{GPU Resources} 
All experiments were conducted on a single NVIDIA RTX PRO 6000 Blackwell Max-Q GPU (96GB), using bfloat16 precision for inference without quantization.

\paragraph{Option-based Tasks}
These tasks require the model to select an answer from a given set of candidates. This category includes the general knowledge and commonsense reasoning tasks described in Appendix~\ref{app:benchmarks} Given an input prompt, we compute the probability of each candidate answer from the final logits and select the candidate with the highest probability as the model prediction. This allows deterministic evaluation.

\paragraph{Generation-based Tasks}
These tasks require the model to generate an output sequence, which is then filtered and compared against the reference answer. This category includes the mathematical reasoning, code generation, and instruction-following tasks described in Appendix~\ref{app:benchmarks}. We use greedy decoding for deterministic evaluation and follow the task-specific filtering and answer extraction procedures provided by \texttt{lm\_eval} framework.

\paragraph{top-$\boldsymbol{p}$ Evaluation}
For top-$p$ routing, we sweep the threshold over $p \in \{0.1, 0.2, \ldots, 0.9\}$. 
In the main results, $p^*$ denotes the threshold that achieves the best average performance for each task among the evaluated values.

\begin{table*}[t]
    \centering
    
    \resizebox{\linewidth}{!}{
    \begin{tabular}{l|c|c c|c c c c c c c c c}
        \toprule
        \multicolumn{2}{l}{}
        & \multicolumn{1}{c}{}
        & \multicolumn{1}{c}{}
        & \multicolumn{9}{c}{General Knowledge \& Commonsense Reasoning} \\

        \cmidrule(lr){5-13}
        
        \multicolumn{2}{l}{{\textbf{Method}}}
        & \multicolumn{1}{c}{\textbf{w/ LDA}}
        & \multicolumn{1}{c}{\textbf{Avg. $k$}}
        & \textbf{MMLU} & \textbf{Hella.} & \textbf{Wino.} & \textbf{ARC-E} & \textbf{ARC-C} & \textbf{CQA} & \textbf{SciQ} & \textbf{PIQA} & \textbf{Average} \\
        
        \midrule
        \multicolumn{2}{l|}{Baseline} & \xmark & 1+4.00 & 70.67\% & 61.06\% & 73.71\% & 82.41\% & 55.46\% & 72.32\% & 96.60\% & 80.09\% & 74.04\% \\
        
        \cline{1-13}
        \multirow{2}{*}{top-$k$}
        & \multirow{2}{*}{$k$=1+1}
        & \rule{0pt}{2.6ex}\xmark & 1+1.00 & 36.95\% & 40.99\% & 52.80\% & 51.09\% & 27.65\% & 24.41\% & 76.90\% & 65.72\% & 47.06\% \\
        & & \hl{\cmark} & \hl{1+1.00} & \hl{58.08\%} & \hl{48.93\%} & \hl{60.77\%} & \hl{71.00\%} & \hl{40.53\%} & \hl{55.12\%} & \hl{91.30\%} & \hl{73.88\%} & \hl{62.45\%} \\
        
        \cline{1-13}
        \multirow{2}{*}{top-$p$}
        & \multirow{2}{*}{$p$=0.1}
        & \rule{0pt}{2.6ex}\xmark & 1+1.42 & 48.43\% & 47.76\% & 57.45\% & 62.45\% & 36.51\% & 35.87\% & 87.50\% & 69.96\% & 55.74\% \\
        & & \hl{\cmark} & \hl{1+1.41} & \hl{64.33\%} & \hl{52.23\%} & \hl{66.38\%} & \hl{77.57\%} & \hl{47.69\%} & \hl{67.56\%} & \hl{95.60\%} & \hl{76.80\%} & \hl{68.52\%} \\

        \cline{1-13}
        \multicolumn{2}{l|}{\multirow{2}{*}{PESF ($\alpha$=0.7)}}
        & \rule{0pt}{2.6ex}\xmark & 1+3.56 & 63.85\% & 57.60\% & 70.00\% & 79.12\% & 50.08\% & 64.29\% & 90.60\% & 77.09\% & 69.07\% \\
        \multicolumn{2}{l|}{} & \hl{\cmark} & \hl{1+3.57} & \hl{64.70\%} & \hl{57.87\%} & \hl{70.71\%} & \hl{80.59\%} & \hl{50.68\%} & \hl{67.89\%} & \hl{93.70\%} & \hl{78.12\%} & \hl{70.53\%} \\

        \bottomrule
    \end{tabular}}

    \vspace{0.5em}
    
    \resizebox{\linewidth}{!}{
    \begin{tabular}{l|c|cc|ccc ccc c}
        \toprule
        \multicolumn{2}{l}{}
        & \multicolumn{1}{c}{}
        & \multicolumn{1}{c}{}
        & \multicolumn{3}{c}{Mathematical Reasoning}
        & \multicolumn{3}{c}{Code Generation}
        & Instruction-Following \\

        \cmidrule(lr){5-7}
        \cmidrule(lr){8-10}
        \cmidrule(lr){11-11}
        
        \multicolumn{2}{l}{\textbf{Method}}
        & \multicolumn{1}{c}{\textbf{w/ LDA}}
        & \multicolumn{1}{c}{\textbf{Avg. $k$}}
        & \textbf{GSM8K} & \textbf{MATH} & \textbf{Average}
        & \textbf{MBPP} & \textbf{HumanEval} & \textbf{Average}
        & \textbf{IFEval} \\

        \midrule
        \multicolumn{2}{l|}{Baseline} & \xmark & 1+4.00 & 84.15\% & 19.40\% & 51.77\% & 40.80\% & 75.61\% & 58.21\% & 48.43\% \\
        
        \cline{1-11}
        \multirow{2}{*}{top-$k$}
        & \multirow{2}{*}{$k$=1+1}
        & \rule{0pt}{2.6ex}\xmark & 1+1.00 & 01.97\% & 00.00\% & 00.99\% & 00.80\% & 01.22\% & 02.01\% & 09.06\% \\
        & & \hl{\cmark} & \hl{1+1.00} & \hl{63.76\%} & \hl{00.00\%} & \hl{31.88\%} & \hl{35.60\%} & \hl{32.93\%} & \hl{34.27\%} & \hl{23.29\%} \\

        \cline{1-11}
        \multirow{2}{*}{top-$p$}
        & \multirow{2}{*}{$p$=0.1}
        & \rule{0pt}{2.6ex}\xmark & 1+2.09 & 21.53\% & 15.20\% & 18.37\% & 19.60\% & 32.92\% & 26.26\% & 20.51\% \\
        & & \hl{\cmark} & \hl{1+2.07} & \hl{70.73\%} & \hl{40.80\%} & \hl{55.77\%} & \hl{47.80\%} & \hl{57.32\%} & \hl{52.56\%} & \hl{33.09\%} \\

        \cline{1-11}
        \multicolumn{2}{l|}{\multirow{2}{*}{PESF ($\alpha$=0.7)}}
        & \rule{0pt}{2.6ex}\xmark & 1+3.41 & 83.85\% & 1.00\% & 42.43\% & 46.00\% & 54.87\% & 50.44\% & 35.48\% \\
        \multicolumn{2}{l|}{} & \hl{\cmark} & \hl{1+3.41} & \hl{83.39\%} & \hl{7.80\%} & \hl{45.60\%} & \hl{52.60\%} & \hl{56.34\%} & \hl{54.57\%} & \hl{37.70\%} \\
        
        \bottomrule
    \end{tabular}}
    
    \caption{\textbf{Performance across a wide range of downstream tasks under different routing strategies on GLM-4.7-Flash.} We use deterministic algorithms for the evaluations. Avg. $k$ denotes the average number of activated experts.}
    \label{tab:glm4_performance}
\end{table*}
\begin{table*}[ht]
    \centering
    
    \resizebox{\linewidth}{!}{
    \begin{tabular}{l|c c c c c c c c c c c c c c}
        \toprule
        \multicolumn{1}{l}{\textbf{Model}} & \textbf{w/ LDA} & \rule{0pt}{2.6ex}\textbf{MMLU} & \textbf{Hella.} & \textbf{Wino.} & \textbf{ARC-E} & \textbf{ARC-C} & \textbf{CQA} & \textbf{SciQ} & \textbf{PIQA} & \textbf{GSM8K} & \textbf{MATH} & \textbf{MBPP} & \textbf{HumanEval} & \textbf{IFEval}\\
        
        \midrule
        \multirow{2}{*}{\textbf{Qwen3-30B-A3B}}
        & \xmark & 0.8 & 0.8 & 0.7 & 0.6 & 0.7 & 0.5 & 0.7 & 0.8 & 0.7 & 0.8 & 0.7 & 0.8 & 0.4 \\
        & \cmark & 0.8 & 0.8 & 0.6 & 0.5 & 0.5 & 0.4 & 0.7 & 0.7 & 0.3 & 0.7 & 0.7 & 0.7 & 0.3 \\

        \midrule
        \multirow{2}{*}{\textbf{kanana-2-30b-a3b-instruct}}
        & \xmark & 0.7 & 0.7 & 0.7 & 0.8 & 0.7 & 0.7 & 0.5 & 0.5 & 0.6 & 0.8 & 0.5 & 0.6 & 0.6 \\
        & \cmark & 0.6 & 0.8 & 0.3 & 0.6 & 0.8 & 0.8 & 0.4 & 0.6 & 0.4 & 0.4 & 0.5 & 0.6 & 0.5 \\
        
        \bottomrule
    \end{tabular}}
    \caption{\textbf{Selected $\boldsymbol{p}$* values across models and downstream tasks.} $p$* denotes the best-performing threshold among $p \in \{0.1, 0.2, \ldots, 0.9\}$ under top-$p$ routing.}
    \label{tab:optimal_top-p}
    
\end{table*}

%%%%%%%%%%%%%%%%%%%%%%%%%%%%%%%%%%%%%%%%
%%%%% Statistical Significance Test
%%%%%%%%%%%%%%%%%%%%%%%%%%%%%%%%%%%%%%%%
\section{Statistical Significance Test} \label{app:statistical_significance_test}

We use McNemar's test~\citep{McNemars_test} for option-based tasks and paired $t$-test~\citep{paired_t-test} for generation-based tasks. 

\paragraph{McNemar's Test}
For option-based tasks, predictions are obtained deterministically from the same set of examples. Thus, we use McNemar's test to compare paired correctness between reduced routing with and without LDA. Let $A$ denote the event that an example is correct only with LDA, and $B$ denote the event that an example is correct only without LDA. The null hypothesis is defined as
\begin{equation*}
    H_0: P(A) = P(B).
\end{equation*}
This test examines whether the number of examples corrected only by LDA is statistically different from the number of examples correctly predicted only by the baseline.

\paragraph{Paired $t$-test}
For generation-based tasks, we evaluate 5 random seeds under the default generation configuration and compare task-level performance using a paired $t$-test. Let $X_{\text{w/ LDA}}$ and $X_{\text{w/o LDA}}$ denote the performance under the same evaluation setting, The null hypothesis is defined as
\begin{equation*}
    H_0: \mathbb{E}[X_{\text{w/ LDA}} - X_{\text{w/o LDA}}] = 0.
\end{equation*}
This test examines whether the mean performance difference induced by LDA is statistically different from zero.

%%%%%%%%%%%%%%%%%%%%%%%%%%%%%%%%%%%%%%%%
%%%%% Performance on Other Models
%%%%%%%%%%%%%%%%%%%%%%%%%%%%%%%%%%%%%%%%
\section{Performance on Other Models} \label{app:other_performance}

We provide additional results on kanana-2-30b-a3b-instruct and GLM-4.7-Flash to examine whether the effectiveness of LDA is limited to Qwen3-30B-A3B. 
As described in Appendix~\ref{app:models}, these models differ from Qwen3-30B-A3B in their expert configurations, use of shared experts, and gating functions. 
Overall, the results show a similar tendency to the main results in Section~\ref{sec:main_results}: reduced expert activation degrades vanilla routing performance, while LDA mitigates this degradation under the different routing strategies.

% In this section, we experiment on additional models using different SMoE architectures to show that the experimental results of Qwen3-30B-A3B presented in the main text are not limited to specific models. Both kanana-2-30b-a3b-instruct and GLM-4.7-Flash model adopt shared expert configurations in their architectures, showing that our method can be applied across diverse SMoE architectures.

\begin{table*}[t]
    \centering
    \resizebox{\linewidth}{!}{
    \begin{tabular}{l cccc cccc cccc}
        \toprule
        \multicolumn{1}{l}{} & \multicolumn{4}{c}{\textbf{MMLU}} & \multicolumn{4}{c}{\textbf{GSM8K}} & \multicolumn{4}{c}{\textbf{MBPP}} \\
        
        \cmidrule(lr){2-5}
        \cmidrule(lr){6-9}
        \cmidrule(lr){10-13}
        
        \multicolumn{1}{l}{\textbf{Method}} & $k$=8 & $k$=6 & $k$=4 & \multicolumn{1}{c}{$k$=2} & $k$=8 & $k$=6 & $k$=4 & \multicolumn{1}{c}{$k$=2} & $k$=8 & $k$=6 & $k$=4 & \multicolumn{1}{c}{$k$=2} \\
        
        \midrule
        top-$k$ & 77.87\% & 76.92\% & 70.72\% & 27.97\% & 88.02\% & 86.81\% & 74.15\% & 00.99\% & 72.60\% & 68.20\% & 34.80\% & 00.00\% \\
        top-$k$ w/ ECS & 77.87\% & 72.72\% & 24.69\% & 24.56\% & 88.02\% & 77.71\% & 01.06\% & 00.00\% & 72.60\% & 54.60\% & 00.00\% & 00.00\% \\
        top-$k$ w/ RMSS & 77.87\% & 77.30\% & 75.33\% & 65.17\% & 88.02\% & 87.94\% & 86.65\% & 66.11\% & 72.60\% & 70.60\% & 67.60\% & 42.00\% \\

        \midrule
        \hl{\textbf{top-$k$ w/ LDA}} & \hl{\textbf{77.87\%}} & \hl{\textbf{77.34\%}} & \hl{\textbf{75.56\%}} & \hl{\textbf{67.40\%}} & \hl{\textbf{88.02\%}} & \hl{\textbf{88.63\%}} & \hl{\textbf{87.49\%}} & \hl{\textbf{68.54\%}} & \hl{\textbf{72.60\%}} & \hl{\textbf{70.60\%}} & \hl{\textbf{68.00\%}} & \hl{\textbf{44.40\%}} \\
        
        \bottomrule
    \end{tabular}}

    \vspace{0.5em}
    
    \resizebox{\linewidth}{!}{
    \begin{tabular}{l c c c c c c c c c c c c}
        \toprule
        \multicolumn{1}{l}{} & \multicolumn{8}{c}{General Knowledge \& Commonsense Reasoning} & \multicolumn{2}{c}{Mathematical Reasoning} & \multicolumn{2}{c}{Code Generation} \\

        \cmidrule(lr){2-9}
        \cmidrule(lr){10-11}
        \cmidrule(lr){12-13}

        \multicolumn{1}{l}{\textbf{Method}} & \textbf{MMLU} & \textbf{Hella.} & \textbf{Wino.} & \textbf{ARC-E} & \textbf{ARC-C} & \textbf{CQA} & \textbf{SciQ} & \textbf{PIQA} & \textbf{GSM8K} & \textbf{MATH} & \textbf{MBPP} & \textbf{HumanEval} \\
        
        \midrule
        top-$2$ & 27.97\% & 36.02\% & 50.99\% & 41.29\% & 24.83\% & 22.36\% & 70.90\% & 61.48\% & 00.99\% & 00.00\% & 00.00\% & 00.61\% \\
        top-$2$ w/ RMSS \; & 65.17\% & 49.05\% & 57.72\% & 67.00\% & 36.17\% & 63.88\% & 93.10\% & 72.41\% & 66.11\% & 37.00\% & 42.00\% & 26.21\% \\

        \midrule
        \hl{\textbf{top-$2$ w/ LDA}} & \hl{\textbf{67.40\%}} & \hl{\textbf{49.43\%}} & \hl{\textbf{58.41\%}} & \hl{\textbf{71.42\%}} & \hl{\textbf{40.27\%}} & \hl{\textbf{66.17\%}} & \hl{\textbf{93.70\%}} & \hl{\textbf{74.97\%}} & \hl{\textbf{68.54\%}} & \hl{\textbf{42.60\%}} & \hl{\textbf{44.40\%}} & \hl{\textbf{36.59\%}} \\

        \bottomrule
    \end{tabular}}
    \caption{\textbf{Comparison with simple scaling baselines on Qwen3-30B-A3B.} ECS, RMSS, and LDA denote Expert-Count Scaling, RMS Scaling, and Layer-wise Distribution Alignment, respectively. We use deterministic algorithms for the evaluations. LDA generally outperforms simple scaling baselines under reduced top-$k$ routing.}
    \label{tab:qwen3_global_scaling_baselines}
\end{table*}

\subsection{kanana-2-30b-a3b-instruct}
Table~\ref{tab:kanana2_performance} reports the results on kanana-2-30b-a3b-instruct. 
Under fixed top-$k$ routing, performance decreases as the number of activated experts is reduced, and the degradation becomes more pronounced in smaller $k$ settings. However, applying LDA consistently improves performance over the corresponding reduced top-$k$ baseline across all task categories.

Under the top-$p$ routing, lowering the threshold $p$ reduces the average number of activated experts and degrades vanilla performance, whereas LDA improves performance under the same top-$p$ setting. In the top-$p$* setting, LDA achieves higher performance than the default top-$k_0$ routing baseline across diverse tasks, indicating that LDA also remains effective under dynamic routing.

We also observe consistent improvements when LDA is combined with PESF. This suggests that LDA is not limited to fixed reduced top-$k$ or dynamic top-$p$ routing, but can also mitigate performance degradation under dynamic expert pruning.

\subsection{GLM-4.7-Flash}
Table~\ref{tab:glm4_performance} reports the results on GLM-4.7-Flash. We observe a similar pattern to the other models. Under aggressive reduced top-$k$ routing, vanilla performance drops severely, whereas LDA substantially mitigates this degradation under the same expert budget. A similar trend is observed under dynamic top-$p$ routing and PESF, where LDA improves performance over the corresponding routing baseline in most task categories.

These additional results suggest that LDA is not restricted to a single SMoE architecture. Across different model architectures and routing strategies, LDA consistently improves reduced-routing performance, supporting the importance of maintaining representation scale and distribution consistency under reduced expert activation.

%%%%%%%%%%%%%%%%%%%%%%%%%%%%%%%%%%%%%%%%%%%%%%%
%%%%% Comparison with Simple Scaling Baselines
%%%%%%%%%%%%%%%%%%%%%%%%%%%%%%%%%%%%%%%%%%%%%%%
\section{Comparison with Simple Scaling Baselines} \label{app:simple_scaling_baselines}

To examine whether the effect of LDA can be explained solely by scalar rescaling of SMoE outputs, we compare LDA with two simple scaling baselines: Expert-Count Scaling and RMS Scaling. These baselines are training-free and applied at inference time, similar to LDA. However, unlike LDA, they apply a single scalar correction to the entire SMoE output and do not align per-dimension distribution statistics. Thus, this comparison allows us to examine whether global scale correction is sufficient, or whether per-dimension distribution alignment is necessary for preserving downstream performance under reduced expert activation.

\paragraph{Expert-Count Scaling}
Expert-Count Scaling is a naive scaling baseline that rescales the SMoE output according to the ratio between the default and reduced numbers of activated experts. 
Given the SMoE output $\mathbf{y}^{(l)}$ at layer $l$, it applies
\[
\hat{\mathbf{y}}^{(l)}
=
\mathbf{y}^{(l)} \cdot \frac{k_0}{k},
\]
where $k_0$ is the default number of activated experts and $k$ is the reduced number of activated experts. This baseline tests whether performance degradation under reduced expert activation can be mitigated by a simple multiplier based only on the expert count.

\paragraph{RMS Scaling}
RMS Scaling rescales the SMoE output using the average RMS scale estimated for each layer and top-$k$ setting. Given the SMoE output $\mathbf{y}^{(l)}$ at layer $l$, it applies
\[
\hat{\mathbf{y}}^{(l)}
=
\mathbf{y}^{(l)} \cdot \frac{\rho^{(l)}_{k_0}}{\rho^{(l)}_k},
\]
where $\rho^{(l)}_k = \mathbb{E}[\mathrm{RMS}(\mathbf{y}_{k}^{(l)})]$ denotes the average RMS of SMoE outputs at layer $l$ under top-$k$ routing. Unlike Expert-Count Scaling, RMS Scaling uses calibration statistics to directly correct the average output scale. However, it still applies a single scalar correction and does not account for dimension-wise distributional mismatch.

\par\medskip
Table~\ref{tab:qwen3_global_scaling_baselines} compares LDA with simple scaling baselines. Expert-Count Scaling leads to substantial performance degradation compared to the vanilla reduced top-$k$ baseline. This indicates that the performance drop under reduced expert activation cannot be addressed by a naive expert-count-based multiplier. Since routing scores are re-normalized, the effect of reducing $k$ is not simply proportional to the number of activated experts.

RMS Scaling provides a stronger baseline. Compared to the vanilla reduced top-$k$ baseline, RMS Scaling substantially mitigates performance degradation across tasks, especially in low-$k$ settings. This result supports our analysis that the scale of SMoE outputs plays an important role in maintaining downstream performance under reduced expert activation.

However, LDA generally achieves higher performance than the simple scaling baselines. Unlike Expert-Count Scaling and RMS Scaling which apply a single scalar correction to the entire SMoE output, LDA aligns the per-dimension mean and standard deviation of reduced top-$k$ representations to those of the default top-$k_0$ configuration. The consistent improvement of LDA over the simple scaling baselines suggests that the mismatch induced by reduced expert activation is not only a global scale shift, but also involves dimension-wise distributional changes. These results empirically provide additional evidence that per-dimension distribution alignment is more effective than global scaling for mitigating routing-induced representation distribution mismatch.
% \paragraph{Per-dimension Variance Alignment}

%%%%%%%%%%%%%%%%%%%%%%%%%%%%%%%%%%%%%%%%
%%%%% Ablation Study
%%%%%%%%%%%%%%%%%%%%%%%%%%%%%%%%%%%%%%%%
\section{Ablation Study} \label{app:ablation_study}

LDA estimates layer-wise distribution statistics from a calibration dataset and uses them to align SMoE outputs under reduced expert activation. Since the correction is determined by calibration statistics, its effectiveness may depend on the choice of calibration dataset and the number of calibration samples. We therefore conduct ablation studies to examine the robustness of LDA with respect to these factors.

However, our calibration procedure relies on activation-level statistics of SMoE outputs, similar to post-training compression methods. 
Prior work on activation-aware quantization and activation-based pruning has shown that activation-based statistics can be estimated effectively from relatively small calibration sets~\citep{wandb} and are often robust across calibration data choices~\citep{awq}. 
This suggests that the role of calibration data is primarily to provide stable estimates of the general representation distribution, rather than to encode task-specific knowledge.

Nevertheless, because LDA explicitly uses calibration statistics for inference-time correction, we empirically examine whether calibration dataset choice and calibration sample size affect downstream performance. We first compare calibration datasets from different domains and then vary the number of calibration samples used to estimate the layer-wise statistics. These ablations provide additional evidence that LDA remains stable across calibration settings.

\begin{table}[t]
    \centering
    \resizebox{\columnwidth}{!}{
    \begin{tabular}{l cc cc cc}
        \toprule
        \multicolumn{1}{l}{\multirow{2}{*}[-1.2ex]{\textbf{\shortstack[l]{Calibration\\Dataset}}}} & \multicolumn{2}{c}{\textbf{MMLU}} & \multicolumn{2}{c}{\textbf{GSM8K}} & \multicolumn{2}{c}{\textbf{MBPP}} \\

        \cmidrule(lr){2-3}
        \cmidrule(lr){4-5}
        \cmidrule(lr){6-7}
        
        \multicolumn{1}{l}{} & $k$=6 & $k$=4 & $k$=6 & $k$=4 & $k$=6 & $k$=4 \\
        
        \midrule
        C4 & 77.34\% & 75.56\% & 88.63\% & 87.49\% & 70.60\% & 68.00\% \\
        GSM8K & 77.28\% & 75.52\% & 88.10\% & 87.41\% & 70.60\% & 68.40\% \\
        MATH & 77.32\% & 75.59\% & 88.09\% & 87.26\% & 70.00\% & 67.80\% \\
        MBPP & 77.35\% & 75.57\% & 88.09\% & 86.65\% & 71.00\% & 67.60\% \\
        
        \bottomrule
    \end{tabular}}
    \caption{\textbf{Effect of calibration dataset choice on Qwen3-30B-A3B.} All settings use top-$k$ routing with LDA and 4$\times$2,048 calibration tokens for layer-wise distribution estimation. We use deterministic algorithms for the evaluations.}
    \label{tab:ablation-qwen3_calibration_dataset}
\end{table}
\begin{table}[t]
    \centering
    \resizebox{\columnwidth}{!}{
    \begin{tabular}{l cc cc cc}
        \toprule
        \multicolumn{1}{l}{} & \multicolumn{2}{c}{\textbf{MMLU}} & \multicolumn{2}{c}{\textbf{GSM8K}} & \multicolumn{2}{c}{\textbf{MBPP}} \\
        
        \cmidrule(lr){2-3}
        \cmidrule(lr){4-5}
        \cmidrule(lr){6-7}
        
         \multicolumn{1}{l}{\textbf{\# of tokens}} & $k$=6 & $k$=4 & $k$=6 & $k$=4 & $k$=6 & $k$=4 \\
        
        \midrule
        1$\times$2048 & 77.39\% & 75.67\% & 88.32\% & 87.02\% & 70.00\% & 67.60\% \\
        2$\times$2048 & 77.29\% & 75.69\% & 88.55\% & 87.26\% & 70.40\% & 67.00\% \\
        4$\times$2048 & 77.34\% & 75.56\% & 88.63\% & 87.49\% & 70.60\% & 68.00\% \\
        8$\times$2048 & 77.35\% & 75.52\% & 88.63\% & 87.71\% & 70.60\% & 67.00\% \\

        \bottomrule
    \end{tabular}}
    \caption{\textbf{Effect of calibration sample size on Qwen3-30B-A3B.} All settings use top-$k$ routing with LDA and C4 as the calibration dataset. We use deterministic algorithms for the evaluations.}
    \label{tab:ablation-qwen3_num_samples}
\end{table}

\subsection{Effect of Calibration Dataset Choice} \label{app:ablation-calibration_dataset}

\begin{table*}[t]
    \centering
    
    \resizebox{\linewidth}{!}{
    \begin{tabular}{l ccc ccc c ccc ccc c}
        \toprule
        \multicolumn{1}{l}{} & \multicolumn{7}{c}{\textbf{kanana-2-30b-a3b-instruct (Layer 24)}} & \multicolumn{7}{c}{\textbf{GLM-4.7-Flash (Layer 24)}} \\

        \cmidrule(lr){2-8}
        \cmidrule(lr){9-15}
        
        \multicolumn{1}{l}{} & \multicolumn{3}{c}{Variance} & \multicolumn{3}{c}{Scale Ratio} & Routing Match Rate & \multicolumn{3}{c}{Variance} & \multicolumn{3}{c}{Scale Ratio} & Routing Match Rate \\
        
        \cmidrule(lr){2-4}
        \cmidrule(lr){5-7}
        \cmidrule(lr){8-8}
        \cmidrule(lr){9-11}
        \cmidrule(lr){12-14}
        \cmidrule(lr){15-15}
        
        \multicolumn{1}{l}{\textbf{Method}} & $k$=2+6 & $k$=2+4 & $k$=2+2 & $k$=2+6 & $k$=2+4 & $k$=2+2 & $k$=2+2 & $k$=1+4 & $k$=1+2 & $k$=1+1 & $k$=1+4 & $k$=1+2 & $k$=1+1 & $k$=1+1 \\
        
        \midrule
        top-$k$ & 0.0179 & 0.0260 & 0.0508 & 0.2180 & 0.2626 & 0.3662 & 62.08\% & 0.0046 & 0.0071 & 0.0121 & 0.2954 & 0.3652 & 0.4764 & 58.93\% \\
        top-$k$ w/ LDA & 0.0179 & 0.0179 & 0.0179 & 0.2180 & 0.2181 & 0.2175 & 72.33\% & 0.0046 & 0.0046 & 0.0046 & 0.2954 & 0.2956 & 0.2948 & 80.49\% \\

        \bottomrule
    \end{tabular}}
    
    \caption{\textbf{Effect of LDA on representations in the other evaluated SMoE architectures.} For both kanana-2-30b-a3b-instruct and GLM-4.7-Flash, LDA restores the increased SMoE output variance and average scale ratio, i.e., $\|\text{SMoE output}\| / \|\text{Residual}\|$, under reduced top-$k$ routing toward the default configuration, while improving the average routing match rate (Eq.~\ref{eq:routing_match_rate}). These results show the same overall tendency as observed in Qwen3-30B-A3B.}
    \label{tab:kanana2_glm4_observations}
\end{table*}

We first examine the effect of calibration dataset choice. Table~\ref{tab:ablation-qwen3_calibration_dataset} compares downstream performance when the layer-wise distribution statistics are estimated from different calibration datasets, including C4, GSM8K, MATH, and MBPP.  All settings use the same number of calibration tokens, and apply LDA under reduced top-$k$ routing.

Across different calibration datasets, the performance remains stable. For MMLU, GSM8K, and MBPP, calibration with C4 achieves performance comparable to calibration with task-specific or domain-specific datasets. Although small variations appear across tasks and top-$k$ settings, no calibration dataset consistently outperforms the others. These results indicate that using a dataset from the same downstream domain does not necessarily yield better performance.

This observation supports the role of calibration data in LDA. The calibration dataset is not used to inject task-specific knowledge or adapt the model to a particular domain. Instead, it is used to estimate layer-wise representation statistics under different routing configurations. Therefore, even when calibration datasets differ in domain, they can provide sufficiently stable estimates of the general representation distribution of SMoE outputs.

\subsection{Effect of Calibration Sample Size} \label{app:ablation-num_samples}

We next examine the effect of the number of calibration samples. Table~\ref{tab:ablation-qwen3_num_samples} compares downstream performance when varying the number of calibration tokens used to estimate the layer-wise distribution statistics. We use C4 as the calibration dataset and apply LDA under reduced top-$k$ routing.

Across different calibration sample sizes, the performance remains relatively stable. Increasing the number of calibration tokens does not lead to consistent performance improvements, and even a small number of calibration samples provides competitive results. This indicates that the layer-wise mean and standard deviation of SMoE outputs can be estimated sufficiently well from a small calibration set. In addition, the lower performance variation across calibration sample sizes further suggests that LDA is not overly sensitive to a particular calibration subset and can be applied with a small calibration cost.

%%%%%%%%%%%%%%%%%%%%%%%%%%%%%%%%%%%%%%%%
%%%%% Algorithms
%%%%%%%%%%%%%%%%%%%%%%%%%%%%%%%%%%%%%%%%
\section{Algorithms} \label{app:algorithms}

We provide pseudo-code descriptions of our method. Algorithm~\ref{alg:layer-wise_distribution_estimation} describes the calibration procedure, where layer-wise distribution statistics are estimated for each top-$k$ setting using an MoE model and a calibration dataset. Algorithm~\ref{alg:distribution-consistent_inference} describes the inference procedure, where the estimated layer-wise statistics are applied to align SMoE outputs under reduced expert activation.

%%%%%%%%%%%%%%%%%%%%%%%%%%%%%%%%%%%%%%%%
%%%%% Additional Plots
%%%%%%%%%%%%%%%%%%%%%%%%%%%%%%%%%%%%%%%%
\section{Additional Plots}
\begin{figure*}[t]
    \includegraphics[width=\linewidth]{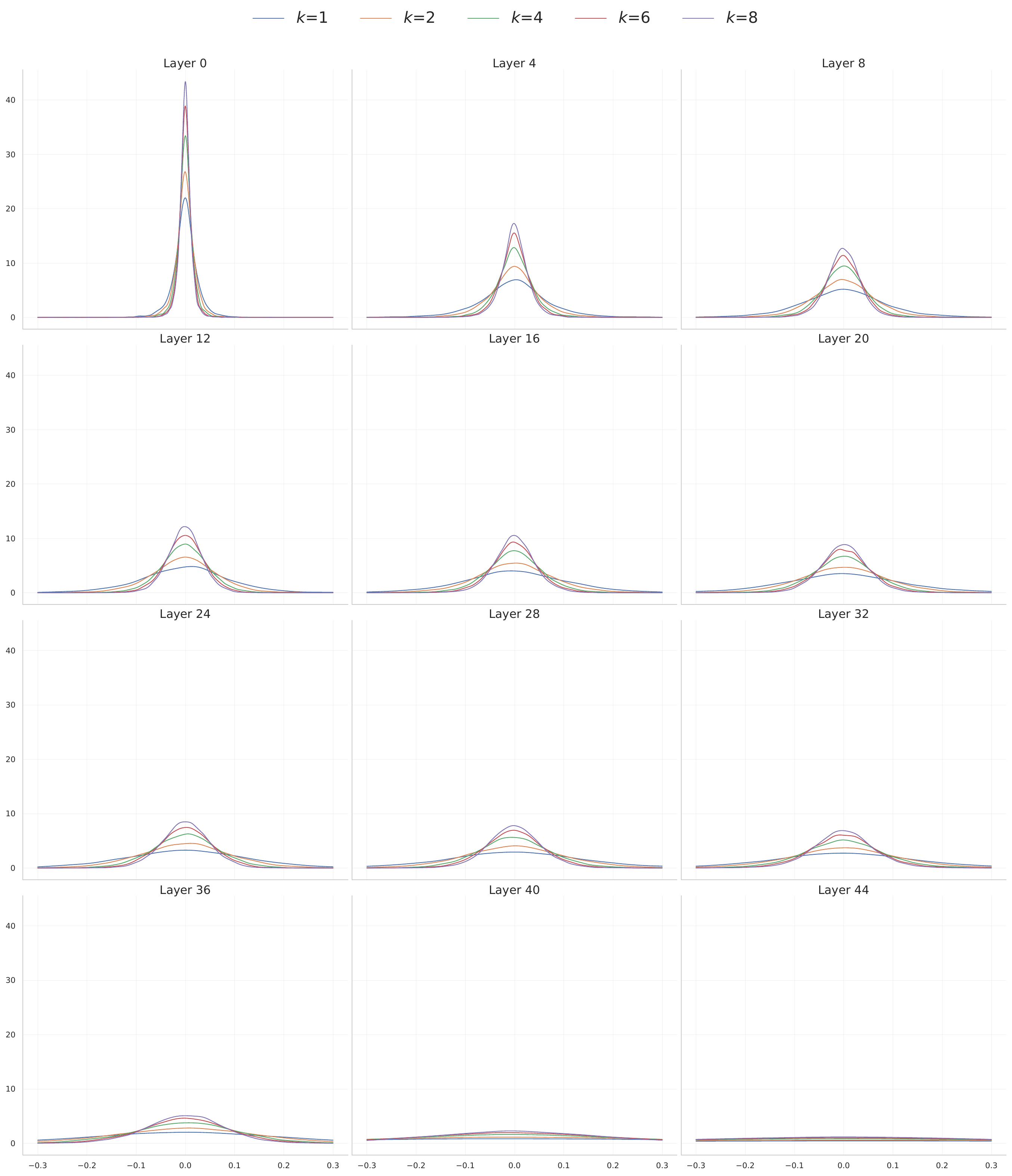}
    \caption{\textbf{Layer-wise distributions of SMoE outputs under different top-$\boldsymbol{k}$ settings before applying LDA on Qwen3-30B-A3B.} We use 2,048 held-out calibration tokens and sample 100,000 values. The distributions become more dispersed under smaller $k$ across layers, consistent with Fig.~\ref{fig:qwen3_observations:a}.}
    \label{fig:qwen3_moe_distribution_all_layer}
\end{figure*}

\begin{figure*}[t]
    \includegraphics[width=\linewidth]{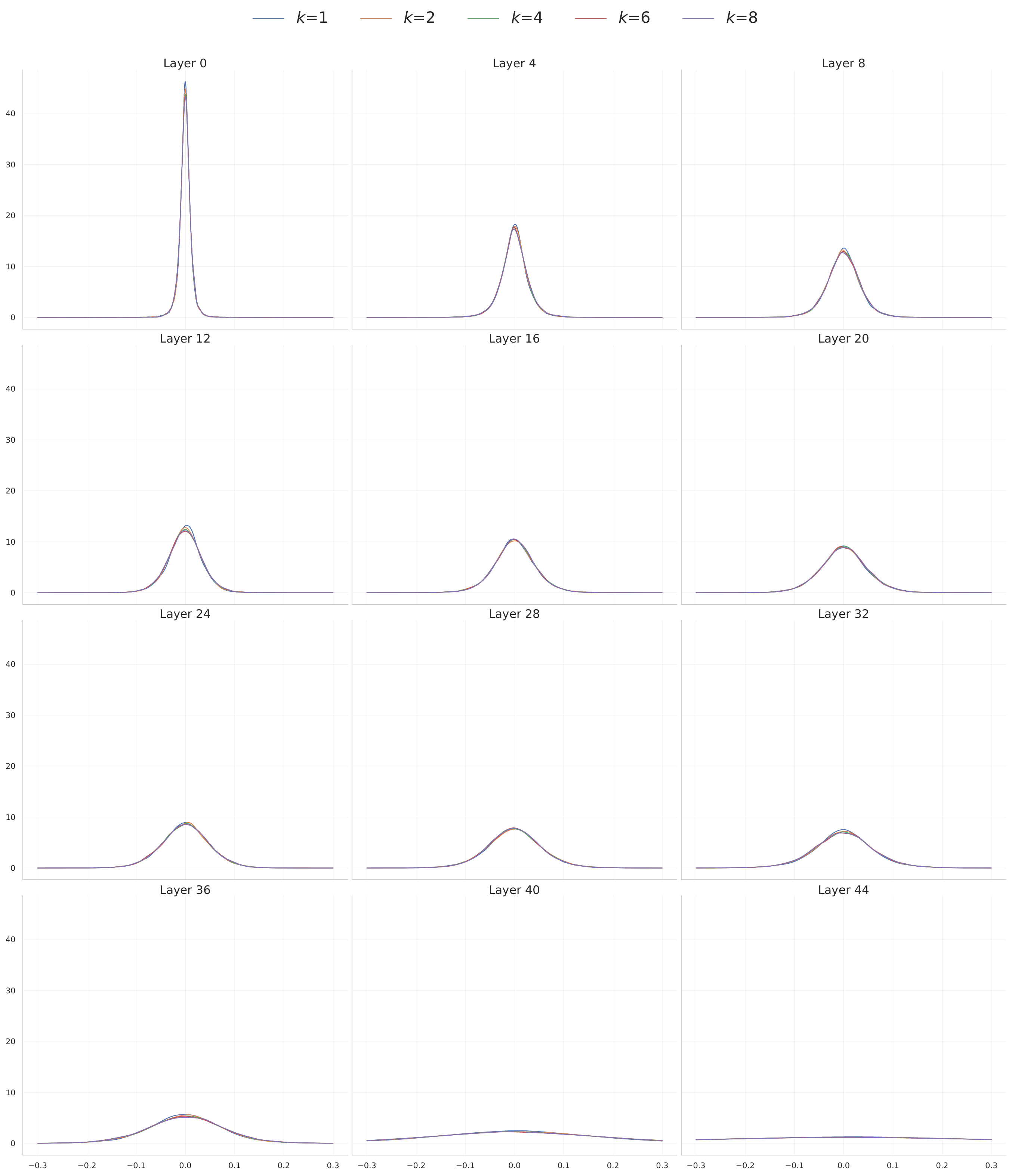}
    \caption{\textbf{Layer-wise distributions of SMoE outputs under different top-$\boldsymbol{k}$ settings after applying LDA on Qwen3-30B-A3B.} We use 2,048 held-out calibration tokens and sample 100,000 values. Compared with Fig.~\ref{fig:qwen3_moe_distribution_all_layer}, the layer-wise distributions become more closely aligned with the default top-$k_0$ setting.}
    \label{fig:qwen3_moe_distribution_all_layer_lda}
\end{figure*}

We provide additional analysis figures for Qwen3-30B-A3B. Figure~\ref{fig:qwen3_moe_distribution_all_layer} shows the layer-wise distributions of SMoE outputs under different top-$k$ settings. Figure~\ref{fig:qwen3_moe_distribution_all_layer_lda} shows the corresponding distributions after applying LDA. These figures provide additional evidence that reduced top-$k$ changes the SMoE output distribution across all layers, while LDA aligns it toward the default top-$k_0$ configuration.

\section*{AI Assistant Usage Statement}
AI assistants were used solely for language polishing, grammar correction, and improving clarity of presentation. All technical content, experimental design, and analysis are entirely our own. 

\newpage
\begin{figure*}[t]
\centering
\begin{minipage}{\linewidth}

\begin{algorithm}[H]
    \centering
    \caption{Layer-wise Distribution Estimation}
    \label{alg:layer-wise_distribution_estimation}
    \begin{algorithmic}[1]
        \STATE \textbf{Input:} calibration dataset $\mathcal{D}$, SMoE model $\mathcal{M}$
        \STATE \textbf{Output:} layer-wise distribution statistics $\mathcal{S}$
        \STATE
        \STATE $L \leftarrow$ number of layers in $\mathcal{M}$
        \STATE $k_0 \leftarrow$ default top-$k$ routing value of $\mathcal{M}$
        \STATE $\mathcal{K} \leftarrow \{1,2,\dots,k_0\}$
        \STATE
        \STATE $\triangleright$ \textbf{Step 1: Sample calibration data}
        \STATE Sample a calibration mini-batch $X \sim \mathcal{D}$
        \STATE
        \STATE $\triangleright$ \textbf{Step 2: Estimate layer-wise distributions}
        \STATE $H^{(0)} \leftarrow \mathrm{Embed}_{\mathcal{M}}(X)$
        \FOR{$l = 0,1,\dots,L-1$}
            \IF{$\mathrm{SMoE} \in \mathrm{Layer}_{\mathcal{M}}^{(l)}$}
                \STATE $\triangleright$ \textbf{Estimate the statistics for each top-$k$ setting}
                \FOR{$k \in \mathcal{K}$}
                    \STATE $Y_k^{(l)} \leftarrow \mathrm{SMoEOutput}^{(l)}_{\mathcal{M}}(H^{(l)}; k)$
                    \STATE $\mu_k^{(l)} \leftarrow \mathrm{Mean}_{\text{token-wise}}\!\left(Y_k^{(l)}\right)$ \hfill $\triangleright$ Per-dimension mean
                    \STATE $\sigma_k^{(l)} \leftarrow \mathrm{Std}_{\text{token-wise}}\!\left(Y_k^{(l)}\right)$ \hfill $\triangleright$ Per-dimension standard deviation
                    \STATE Store $\{\mu_k^{(l)}, \sigma_k^{(l)}\}$ in $\mathcal{S}$
                \ENDFOR
            \ENDIF
            \STATE $\triangleright$ \textbf{Propagate with default top-$k_0$ routing}
            \STATE $H^{(l+1)} \leftarrow \mathrm{Layer}^{(l)}_{\mathcal{M}}(H^{(l)}; k_0)$
        \ENDFOR
        \STATE
        \STATE \textbf{Return} $\mathcal{S}$
    \end{algorithmic}
\end{algorithm}

\end{minipage}
\end{figure*}

\columnbreak
\begin{figure*}[t]
\centering
\begin{minipage}{\linewidth}

\begin{algorithm}[H]
    \caption{Distribution-Consistent Inference}
    \label{alg:distribution-consistent_inference}
    \begin{algorithmic}[1]
        \STATE \textbf{Input:} input data $X$, distribution statistics $\mathcal{S}$, SMoE model $\mathcal{M}$, routing value $k$
        \STATE \textbf{Output:} model output $\hat{X}$
        \STATE
        \STATE $L \leftarrow$ number of layers in $\mathcal{M}$
        \STATE $k_0 \leftarrow$ default top-$k$ routing value of $\mathcal{M}$
        \STATE
        \STATE $H^{(0)} \leftarrow \mathrm{Embed}_{\mathcal{M}}(X)$
        \FOR{$l = 0,1,\dots,L-1$}
            \STATE $\triangleright$ \textbf{Propagate to the next layer}
            \IF{$\mathrm{SMoE} \in \mathrm{Layer}_{\mathcal{M}}^{(l)}$}
                \STATE $\triangleright$ \textbf{Forward with top-$k$ routing}
                \STATE $Y_k^{(l)} \leftarrow \mathrm{SMoEOutput}_{\mathcal{M}}^{(l)}(H^{(l)}; k)$
                \IF{$k < k_0$}
                    \STATE $\triangleright$ \textbf{Apply layer-wise distribution alignment}
                    \STATE $\{\mu_{k_0}^{(l)}, \sigma_{k_0}^{(l)}\} \leftarrow \mathcal{S}[l][k_0]$ \hfill $\triangleright$ Reference statistics
                    \STATE $\{\mu_{k}^{(l)}, \sigma_{k}^{(l)}\} \leftarrow \mathcal{S}[l][k]$ \hfill $\triangleright$ Target statistics
                    \STATE $\hat{Y}^{(l)} \leftarrow \sigma_{k_0}^{(l)} \odot \dfrac{Y_k^{(l)} - \mu_k^{(l)}}{\sigma_k^{(l)} + \epsilon} + \mu_{k_0}^{(l)}$  \hfill $\triangleright$ Per-dimension correction
                \ELSE
                    \STATE $\hat{Y}^{(l)} \leftarrow Y_k^{(l)}$
                \ENDIF
                \STATE $H^{(l+1)} \leftarrow \mathrm{Layer}_{\mathcal{M}}^{(l)}(H^{(l)}; k, \hat{Y}^{(l)})$
            \ELSE
                \STATE $H^{(l+1)} \leftarrow \mathrm{Layer}_{\mathcal{M}}^{(l)}(H^{(l)})$
            \ENDIF
        \ENDFOR
        \STATE
        \STATE $\hat{X} \leftarrow \mathrm{Output}_{\mathcal{M}}(H^{(L)})$
        \STATE \textbf{Return} $\hat{X}$
    \end{algorithmic}
\end{algorithm}

\end{minipage}
\end{figure*}


\newpage
\begin{table*}[t]
    \centering
    \small
    \setlength{\tabcolsep}{5pt}
    \renewcommand{\arraystretch}{1.15}
    \begin{tabularx}{\linewidth}{l Y l}
        \toprule
        \textbf{\normalsize Artifact} & \textbf{\normalsize Description} & \textbf{\normalsize License} \\

        \midrule
        \rule{0pt}{1.0ex}MMLU 
        & General knowledge and reasoning benchmark; English; auxiliary\_train/dev/val/test = 99,842/285/1,531/14,042 samples. 
        & MIT License \\

        \cline{1-3}
        \rule{0pt}{2.5ex}HellaSwag 
        & Commonsense reasoning benchmark; English; train/val/test = 39,905/10,042/10,003 samples. 
        & MIT License \\

        \cline{1-3}
        \rule{0pt}{2.5ex}WinoGrande 
        & Commonsense coreference reasoning benchmark; English; train/val/test = 40,398/1,267/1,767 samples. 
        & CC-BY License \\

        \cline{1-3}
        \rule{0pt}{2.5ex}ARC-Easy 
        & Science question answering benchmark; English; train/val/test = 2,251/570/2,376 samples. 
        & CC BY-SA 4.0 License \\

        \cline{1-3}
        \rule{0pt}{2.5ex}ARC-Challenge 
        & Challenging science question answering benchmark; English; train/val/test = 1,119/299/1,172 samples. 
        & CC BY-SA 4.0 License \\

        \cline{1-3}
        \rule{0pt}{2.5ex}CommonsenseQA 
        & Commonsense question answering benchmark; English; train/val/test = 9,741/1,221/1,140 samples. 
        & MIT License \\

        \cline{1-3}
        \rule{0pt}{2.5ex}SciQ 
        & Science question answering benchmark; English; train/val/test = 11,679/1,000/1,000 samples. 
        & CC BY-NC 2.5 License \\

        \cline{1-3}
        \rule{0pt}{2.5ex}PIQA 
        & Physical commonsense reasoning benchmark; English; train/val/test = 16,113/1,838/3,084 samples. 
        & Academic Free License v2.5 \\

        \cline{1-3}
        \rule{0pt}{2.5ex}GSM8K 
        & Grade-school mathematical reasoning benchmark; English; train/test = 7,473/1,319 samples. 
        & MIT License \\

        \cline{1-3}
        \rule{0pt}{2.5ex}MATH500 
        & Competition-level mathematical reasoning benchmark; English; 500 samples. 
        & MIT License \\

        \cline{1-3}
        \rule{0pt}{2.5ex}MBPP 
        & Python code generation benchmark from natural language descriptions; English/Python; train/val/test = 374/90/500 samples. 
        & CC BY 4.0 License \\

        \cline{1-3}
        \rule{0pt}{2.5ex}HumanEval 
        & Python code generation benchmark; English/Python; 164 samples. 
        & MIT License \\

        \cline{1-3}
        \rule{0pt}{2.5ex}IFEval 
        & Instruction-following benchmark with verifiable constraints; English; 541 samples. 
        & Apache 2.0 License \\

        \cline{1-3}
        \rule{0pt}{2.5ex}Qwen3-30B-A3B 
        & Sparse MoE LLM used for evaluation. 
        & Apache 2.0 License \\

        \cline{1-3}
        \rule{0pt}{2.5ex}kanana-2-30b-a3b-instruct 
        & Sparse MoE LLM used for evaluation. 
        & Kanana License Agreement \\

        \cline{1-3}
        \rule{0pt}{2.5ex}GLM-4.7-Flash 
        & Sparse MoE LLM used for evaluation. 
        & MIT License \\
        
        \cline{1-3}
        \rule{0pt}{2.5ex}lm\_eval 
        & Evaluation framework for language model benchmarks. 
        & MIT License \\

        \cline{1-3}
        \rule{0pt}{2.5ex}vLLM 
        & Inference framework for efficient LLM serving and evaluation. 
        & Apache 2.0 License \\

        \bottomrule
    \end{tabularx}
    \caption{\textbf{Documentation of experimental artifacts.} We report summary of the datasets, models, and frameworks used in our experiments, including descriptions, split sizes, and licenses.}
    \label{tab:artifacts}
\end{table*}

\end{document}